\documentclass[a4paper,fleqn]{cas-sc}
\usepackage{tabularx}
\usepackage{xltabular}
\usepackage{longtable}
\usepackage{url}
\usepackage{caption}
\usepackage[authoryear,longnamesfirst]{natbib}
\usepackage{multirow}
\usepackage{pifont}

\usepackage{xcolor}

\newcommand{\cmark}{\ding{51}}%
\newcommand{\xmark}{\ding{55}}%

\def\tsc#1{\csdef{#1}{\textsc{\lowercase{#1}}\xspace}}
\tsc{WGM}
\tsc{QE}
\tsc{EP}
\tsc{PMS}
\tsc{BEC}
\tsc{DE}

\begin{document}
\let\WriteBookmarks\relax
\def\floatpagepagefraction{1}
\def\textpagefraction{.001}

\shorttitle{Deciphering knee osteoarthritis diagnostic features with explainable artificial intelligence}

\shortauthors{Teoh et~al.}

\title [mode = title]{Deciphering knee osteoarthritis diagnostic features with explainable artificial intelligence: A systematic review}

%
\author[1,2]{Yun Xin Teoh}[type=editor,
                        auid=000,bioid=1,
                        prefix=,
                        orcid=0000-0002-4729-5235]





\credit{Data curation, writing - original draft}

\affiliation[1]{organization={Department of Biomedical Engineering, Faculty of Engineering},
    addressline={Universiti Malaya}, 
    city={Kuala Lumpur},
    postcode={50603}, 
    country={Malaysia}}
\affiliation[2]{organization={LISSI},
    addressline={Université Paris-Est Créteil}, 
    city={Vitry sur Seine},
    postcode={94400}, 
    country={France}}

\author[2]{Alice Othmani}[orcid=0000-0002-3442-0578]
\cormark[1]
\ead{alice.othmani@u-pec.fr}
\credit{Supervision of project, software, writing - review \& editing}

\author[3,4]{Siew Li Goh}[orcid=0000-0001-5898-1196
   ]
\credit{Conceptualization of this study from clinical perspective}

\affiliation[3]{organization={Sports Medicine Unit, Faculty of Medicine},
    addressline={Universiti Malaya}, 
    city={Kuala Lumpur},
    postcode={50603}, 
    country={Malaysia}}
\affiliation[4]{organization={Centre for Epidemiology and Evidence-Based Practice, Faculty of Medicine},
    addressline={Universiti Malaya}, 
    city={Kuala Lumpur},
    postcode={50603}, 
    country={Malaysia}}

\author%
[1]
{Juliana Usman}[orcid=0000-0001-8983-0892]
\credit{Originate idea of study from biomechanics perspective}

\author[1]{Khin Wee Lai}[orcid=0000-0002-8602-0533]
\cormark[1]
\ead{lai.khinwee@um.edu.my}
\credit{Originate idea of study from engineering perspective}


\cortext[cor1]{Corresponding author}


\begin{abstract}
Existing artificial intelligence (AI) models for diagnosing knee osteoarthritis (OA) have faced criticism for their lack of transparency and interpretability, despite achieving medical-expert-like performance. This opacity makes them challenging to trust in clinical practice. Recently, explainable artificial intelligence (XAI) has emerged as a specialized technique that can provide confidence in the model's prediction by revealing how the prediction is derived, thus promoting the use of AI systems in healthcare. This paper presents the first survey of XAI techniques used for knee OA diagnosis. The XAI techniques are discussed from two perspectives: data interpretability and model interpretability. The aim of this paper is to provide valuable insights into XAI's potential towards a more reliable knee OA diagnosis approach and encourage its adoption in clinical practice. 
\end{abstract}


\begin{highlights}
\item A review of explainable artificial intelligence (XAI) techniques for ensuring clinical trustworthiness of AI-aided knee osteoarthritis (OA) diagnosis.
\item An overview of data interpretability approach used in XAI algorithms.
\item A summary of model interpretability techniques for knee OA diagnosis.
\item A comprehensive discussion of the opportunities and open challenges of implementing XAI for knee OA diagnosis.
\end{highlights}


\maketitle

\section{Introduction}
\label{sec:introduction}

Osteoarthritis (OA) is a prevalent degenerative joint disease that affects millions of people worldwide, with the weight-bearing knee joint being particularly susceptible. While radiography is a commonly used diagnostic tool for knee OA, its diagnostic precision is often compromised due to the subjective nature of image interpretation and the perceptual differences among radiologists, which are influenced by their individual knowledge and experience. To enhance the accuracy of OA diagnosis, researchers have also delved into modeling OA using multidimensional data, such as electronic medical records, to encompass a comprehensive range of patient information. This includes demographic, societal, symptomatic, medical history, biomechanical, biochemical, genetic, and behavioral characteristics. Artificial intelligence (AI) models have demonstrated the ability to automate diagnosis and have shown promising results, achieving diagnostic accuracy on par with medical experts using either individual or combined data \citep{Tiulpin2018, Karim2021}. However, the specific impact of each factor within the data and the interrelationships between these factors remain largely unexplored.

Furthermore, there is a growing concern about the lack of transparency and interpretability of AI models in healthcare settings \citep{Hah2021, Lee2022}. The use of AI models in medical data for OA diagnosis shows potential in reducing the subjectivity and variability linked to human interpretation. However, those AI approaches predominantly rely on black-box models, which lack transparency and interpretability \citep{Hah2021, Lee2022}. In contrast to the human reasoning process, which depends on complex cognitive abilities, intuition, and the assimilation of diverse knowledge and experiences to make decisions, AI models make predictions based on the learning outcomes from training datasets. The internal workings of these models remain hidden or unknown, even to their designers. This lack of transparency can engender uncertainty and erode trust among patients and healthcare providers. Additionally, the use of black-box models impedes the development of health mobile applications for disease management \citep{Mrklas2020}. According to a survey conducted by \citet{Mrklas2020}, a significant number of patients and physicians have expressed a strong desire for a visual symptom graph to aid in monitoring their condition.

Explainable AI (XAI) \citep{Giuste2023} offers a potential solution to these concerns by providing a transparent and interpretable framework for automated analysis of radiographic images. XAI algorithms can identify specific regions of interest within the image and provide a clear explanation of the factors that contributed to the final diagnosis \citep{Van2022explainable, Groen2022systematic}. This could help to overcome the limitations of traditional radiographic diagnosis and increase the accuracy and consistency of knee OA diagnosis.

By leveraging XAI, healthcare providers could have a more objective and transparent method for diagnosing knee OA, leading to earlier detection and more timely treatment. XAI could also potentially reduce the need for costly and invasive diagnostic procedures, such as arthroscopy, which are currently used for further evaluation to confirm cartilage lesion.

\subsection{Motivations and research objectives}
In recent years, as XAI gains popularity, numerous survey papers have emerged discussing its application in healthcare settings (Table \ref{tab:summary_reviews_and_surveys_in_XAI_healthcare}). Despite this growing interest, there is still a noticeable lack of comprehensive survey papers that delve into the specific application of XAI for diagnosing knee OA. Furthermore, many existing XAI strategies have been designed with a general-purpose approach and may not fully address the unique clinical concerns and domain-specific knowledge required for accurate diagnosis of knee OA. Therefore, there is a need for specialized XAI frameworks that take into account the specific clinical considerations and incorporate relevant domain knowledge to enhance the application of XAI in diagnosing knee OA effectively. To address this gap, it is crucial to explore different explanation methods and evaluate their effectiveness. By conducting a comprehensive review of the literature on interpretability and explainability of AI models for knee OA diagnosis, we can gain a deeper understanding of these concepts and their potential applications. Such a review will provide valuable insights into how interpretability and explainability can be leveraged to improve AI and machine learning models for knee OA diagnosis. To the best of our knowledge, this paper represents the first survey dedicated to exploring the application of XAI in knee OA diagnosis. In this study, our objectives are threefold:
\begin{itemize}
    \item Evaluation of the current state-of-the-art explainability and interpretability methods for neural networks used in diagnosing knee OA from medical data;
    \item Comparison of the existing knee OA datasets and the performance analysis of different explainability and interpretability methods in AI models; 
    \item Identification of the potential clinical impact of the most promising explainability and interpretability methods by assessing their practicality, scalability, and effectiveness in real-world clinical settings for improving diagnostic accuracy and reducing misdiagnosis rates in knee OA.
\end{itemize}

\begin{table*}[]
\caption{Summary of existing reviews and surveys on the topic of explainable artificial intelligence (XAI) in healthcare applications.}
\label{tab:summary_reviews_and_surveys_in_XAI_healthcare}
\resizebox{\textwidth}{!}{%
\begin{tabular}{cccccccc}
\hline
\multirow{2}{*}{\textbf{Paper}} & \multirow{2}{*}{\textbf{Year}} & \multirow{2}{*}{\textbf{Topic}} & \multicolumn{5}{c}{\textbf{Scope}} \\ \cline{4-8} 
& & & \textbf{XAI} & \textbf{Tabular data} & \textbf{Image data} & \textbf{Disease-specific} & \textbf{Knee OA} \\ \hline
\citet{Tjoa2020} & 2020 & Medical XAI & \cmark & \xmark & \xmark & \xmark & \xmark \\
\citet{Van2022explainable} & 2022 & XAI for medical imaging & \cmark & \xmark & \cmark & \xmark & \xmark \\
\citet{Groen2022systematic} & 2022 & XAI for radiology & \cmark & \xmark & \cmark & \xmark & \xmark \\  
\citet{Loh2022application} & 2022 & XAI for medical applications & \cmark & \cmark & \cmark & \xmark & \xmark \\  
\citet{Chaddad2023} & 2023 & XAI for medical imaging & \cmark & \xmark & \cmark & \xmark & \xmark \\ 
\citet{Nazir2023survey} & 2023 & XAI for medical imaging & \cmark & \xmark & \cmark & \xmark & \xmark \\ 
\citet{Giuste2023} & 2023 & XAI for COVID-19 & \cmark & \xmark & \xmark & \cmark & \xmark \\
\citet{Joyce2023explainable} & 2023 & XAI for mental health & \cmark & \cmark & \xmark & \cmark & \xmark \\ 
\citet{Bharati2023review} & 2023 & XAI for medical applications & \cmark & \cmark & \cmark & \cmark & \xmark \\ 
Our paper & 2023 & XAI for knee OA diagnosis & \cmark & \cmark & \cmark & \cmark & \cmark \\ 
\hline
\end{tabular}
}
\end{table*}

\subsection{Organization of paper}
This review paper is partitioned into nine sections. Firstly, Section \ref{preliminaries_and_fundamental_concepts} introduces the preliminaries and fundamental concepts of XAI. Section \ref{sec:search_strategy_and_eligibility_criteria} describes the study protocol, including the search strategy, as well as the inclusion and exclusion criteria for selecting relevant studies. In Section \ref{sec:general_query_on_knee_OA_assessment}, the word co-occurrence analysis conducted on the included studies is presented. Following that, Section \ref{sec:data_for_knee_OA_assessment} provides detailed information regarding the knee OA data used in the assessment. Section \ref{sec:classification_systems_for_final prediction_of_knee_OA_conditions} outlines the classification systems employed in predictive modeling for knee OA. Moving on, Section \ref{sec:XAI_approaches} introduces an XAI taxonomy and explores various techniques for achieving data and model interpretability. The implications and potential applications of XAI are discussed in Section \ref{sec:discussion}, which also suggests promising avenues for future research. Finally, Section \ref{sec:conclusion} offers a comprehensive conclusion to this study, summarizing its findings and highlighting its contributions to the field of XAI in knee OA assessment.

\section{Preliminaries and fundamental concepts}
\label{preliminaries_and_fundamental_concepts}

\subsection{XAI concepts and frameworks}
Due to rapid development of artificial intelligence and machine learning technologies, it becomes increasingly important to understand how these models make predictions \citep{Buijsman2022}. In this field, the terms "interpretability" and "explainability" are closely related and often used interchangeably \citep{Linardatos2020}, but they do have subtle differences in the context of deep learning. Here's a breakdown of each concept:

\textbf{Interpretability:} Interpretability refers to the ability to understand and make sense of the internal workings of a deep learning model \citep{Ali2023}. It involves gaining insights into how the model processes inputs, makes decisions, and generates outputs. An interpretable model allows humans to examine and comprehend the underlying mechanisms and logic employed by the model to arrive at its predictions or decisions \citep{Murdoch2019definitions}.

\textbf{Explainability:} Explainability, on the other hand, focuses on providing human-understandable explanations for the model's outputs or predictions \citep{Arrieta2020explainable,Ali2023}. It goes beyond mere interpretation and aims to make the decision-making process of the model transparent and understandable to non-experts. Explainable models not only produce accurate predictions but also provide intuitive explanations that can be easily comprehended by end-users or stakeholders.

In summary, while interpretability is primarily concerned with understanding the internal workings of a deep learning model, explainability goes a step further by providing human-understandable explanations for the model's outputs or decisions. Both concepts aim to enhance the transparency and trustworthiness of deep learning models, especially in high-stakes applications such as healthcare, finance, or autonomous systems.

Recently published XAI taxonomies \citep{Linardatos2020,Schwalbe2023,Chaddad2023} propose a conceptual framework for XAI, utilizing four evaluation dimensions to effectively describe the scope and characteristics of the XAI domain. These dimensions include: 
\begin{itemize}
    \item \textbf{Explanation scopes}, which can be divided into local (explaining individual prediction) or global (explaining the whole model) interpretability.
    \item \textbf{Model specificity}, which can be divided into model-specific and model-agnostic interpretability.
    \item \textbf{Interpretation types}, which can be divided into pre-model, intrinsic, post-hoc, and extrinsic interpretability.
    \item \textbf{Explanation forms}, which encompass various ways in which explanations can be presented or communicated.
\end{itemize}
The proposed XAI framework effectively tackles the technical concerns of general AI models. However, it lacks emphasis on the essential aspects of data and problem characteristics required for instilling domain knowledge into AI models. Moreover, it does not adequately consider the specific needs of lay users, such as medical experts \citep{Du2019techniques}. These factors are crucial in ensuring that AI models are not only transparent and interpretable but also capable of effectively utilizing domain-specific information to enhance their performance and relevance in real-world applications. Therefore, \citep{Nauta2023} extend the general XAI framework by incorporating considerations for the type of input data, problem, and task. This extension aims to provide a more comprehensive and practical approach to XAI, catering to the specific needs of various domains and ensuring the successful integration of domain knowledge into AI models.

The realm of interpretability in XAI can be categorized into two distinct groups: perceptive interpretability and interpretability by mathematical structures, as proposed by \citet{Tjoa2020}. Perceptive interpretability methods typically provide immediate interpretations, while methods that offer interpretation via mathematical structures produce outputs that require an additional layer of cognitive processing to reach a human-readable presentation. These taxonomies primarily focus on the transition from black-box models to white-box models, where the inner logic is fully explored and understood. \citet{Ali2023} introduce a novel approach by incorporating gray-box models. These models lie between black-box and white-box models, offering a partial understanding of the underlying mechanisms. By considering this intermediate category, the proposed taxonomy accounts for a broader range of interpretability levels and provides a more nuanced perspective on XAI. Compared to previous studies, their XAI taxonomy incorporates data explainability as an essential aspect to comprehend the datasets used in the AI models. This addition reflects their effort on providing insights into the transparency and interpretability of the data itself, in addition to understanding the model's decision-making process. By considering data explainability, the proposed taxonomy offers a more comprehensive approach in gaining a deeper understanding of AI systems and the role of data in shaping their predictions.

All previously proposed XAI taxonomies offer a structured framework for comprehending and classifying various aspects of XAI approaches and their applications. As highlighted by \citet{Nauta2023}, it is important to recognize that certain explanation methods have the ability to incorporate multiple types of explanations, thereby making the categories of explanation methods non-mutually exclusive.

\begin{figure*}
\centering
\includegraphics[width=\textwidth]{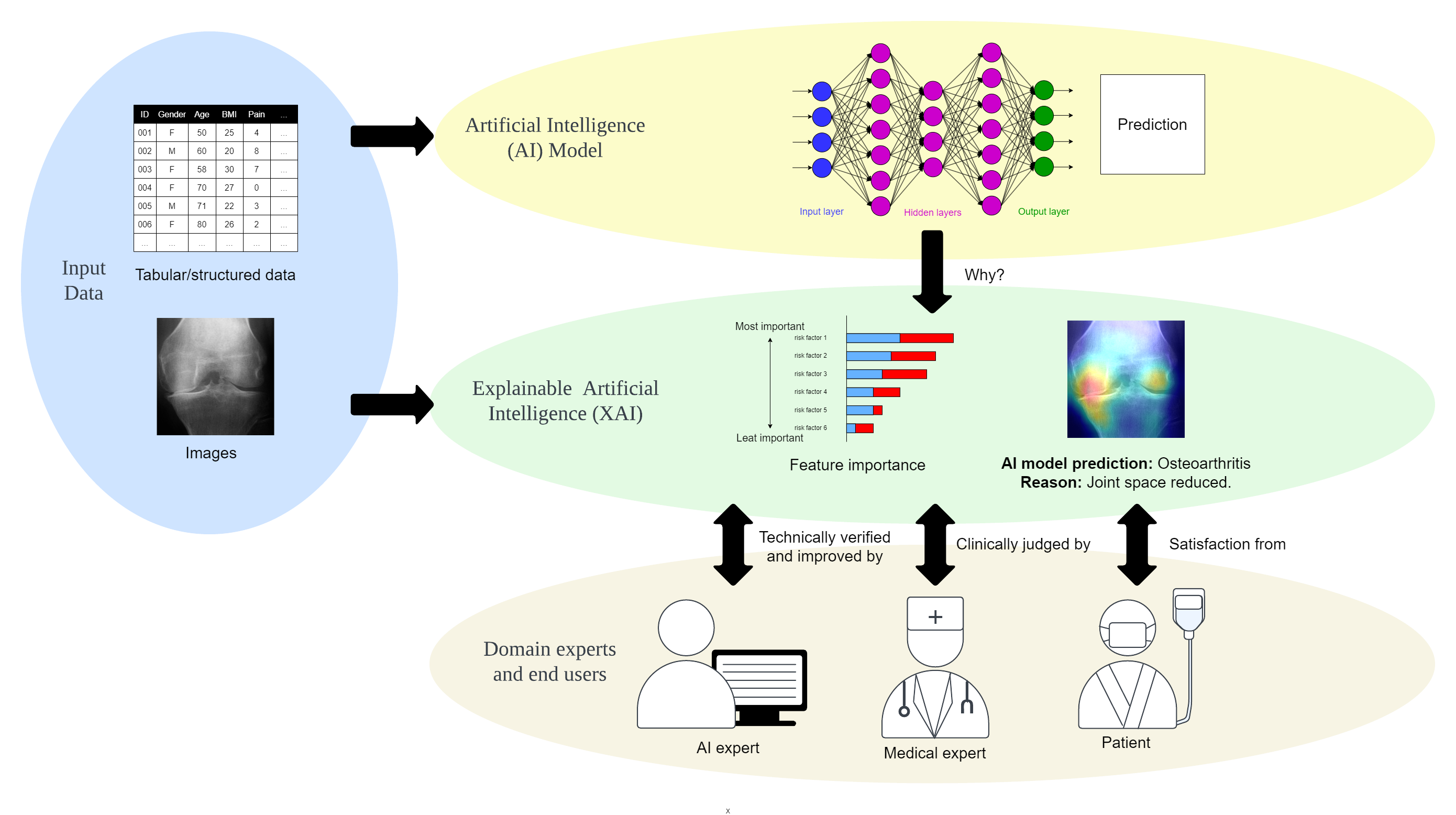}
\caption{Illustration of XAI implementation for knee OA diagnosis. Through XAI, the decision-making process of AI models becomes interpretable and explainable, leading to the visualization of essential insights for AI expert, medical expert, and patient.}\label{fig_user-centric_XAI}
\end{figure*}

In order to enhance the connection between users and XAI, \citet{Wang2019designing} introduced a theoretical conceptual framework that establishes links between different XAI explanation facilities and user reasoning goals. Their work generated a concept called user-centric XAI, where the AI systems are designed by placing the end-users, such as healthcare professionals or patients, at the forefront of the explanation process, as illustrated in \ref{fig_user-centric_XAI}. Their framework was meticulously designed to mitigate reasoning failures caused by cognitive biases. Additionally, \citet{Schoonderwoerd2021} proposed a flowchart to guide the design of human-centered XAI systems. This flowchart incorporates three essential components: domain analysis, requirements analysis, and interaction design. By following this flowchart, XAI designers can ensure that their systems are aligned with user needs and provide effective explanations for improved user understanding and decision-making.

\subsection{Ethical considerations in XAI}
Global policy discussions are placing increasing emphasis on the integration of ethical standards into the design and implementation of AI-enabled technologies, highlighting the growing importance of Trustable AI. In 2018, the High-Level Expert Group on AI, established by the European Commission, published ethical guidelines focused on fostering trust in human-centric AI \citep{AI_Hleg2019ethics}. The guidelines highlighted seven key requirements for Trustable AI \citep{Kumar2020trustworthy}, as follows:
\begin{itemize}
    \item \textbf{Human agency and oversight} that emphasize human autonomy and the importance of fundamental rights in decision-making.
    \item \textbf{Technical robustness and safety} that ensure AI systems are designed to prevent harm and promote resilience and security.
    \item \textbf{Privacy and data governance} that respect privacy and data protection while implementing sound data governance mechanisms.
    \item \textbf{Transparency} that advocates for transparency in data, system, and AI business models, complemented by traceability and explainability.
    \item \textbf{Diversity, non-discrimination, and fairness} that promote fairness and accessibility for all human while involve relevant stakeholders throughout the AI system's lifecycle.
    \item \textbf{Societal and environmental well-being} that focus on AI systems' positive impact on society and the environment, including sustainability considerations.
    \item \textbf{Accountability} that establishes mechanisms for responsibility and accountability, including auditability and accessible redress for AI system outcomes.
\end{itemize}

These requirements lead to the principles of Valid AI, Responsible AI, Privacy-preserving AI, and Explainable AI (XAI):

\begin{itemize}
    \item \textbf{Valid AI} ensures that AI systems produce accurate and reliable results by using high-quality data, appropriate algorithms, and robust evaluation methods. It aims to minimize errors and biases, making the AI outputs valid and trustworthy.
    
    \item \textbf{Responsible AI} involves designing and deploying AI systems in an ethical and socially conscious manner. It entails considering potential societal impacts, adhering to human values, and complying with legal and regulatory standards to minimize harm and promote positive outcomes.
    
    \item \textbf{Privacy-preserving AI} safeguards individuals' sensitive data during data processing and model training. These AI techniques ensure that personal information remains protected and confidential, preventing unauthorized access and preserving user privacy.
    
    \item \textbf{Explainable AI (XAI)} addresses the question of understanding the reasoning behind AI decisions. It provides transparency and interpretability to AI outputs, allowing users, including AI experts, medical professionals, and patients, to comprehend and trust the AI model's decisions.
\end{itemize}

In this framework, XAI plays a crucial role in addressing the fundamental question surrounding the rationale behind the decision-making process of AI systems, encompassing both human-level XAI (for human users) and machine-level XAI (for other AI models or systems). XAI techniques contribute to the transparency and interpretability required for achieving Trustable AI.

The European Union's High-Level Group on AI has made significant efforts to promote XAI by taking initiatives such to implement the General Data Protection Regulation (GDPR) \citep{Mondschein2019,Hamon2022}. In addition, the proposal of the Artificial Intelligence Act by the European Commission represents their recent endeavors to foster a robust internal market for Artificial Intelligence (AI) systems \citep{Kop2021,Van2022}. In the United States, the Defense Advanced Research Projects Agency (DARPA) has launched an XAI program aimed at tackling three key challenges: (1) developing more explainable models, (2) designing effective explanation interfaces, and (3) understanding the psychological requirements for effective explanations \citep{Dw2019}. Despite considerable efforts, existing explainability methods still fall short in providing reassurance about the correctness of individual decisions, building trust among users, and justifying the acceptance of AI recommendations in clinical practice. Consequently, there is an immediate need to prioritize rigorous internal and external validation of AI models as a more direct approach to achieving the goals commonly associated with explainability \citep{Ghassemi2021}.

\section{Search strategy and eligibility criteria}
\label{sec:search_strategy_and_eligibility_criteria}
This systematic review was conducted based on the procedure proposed by \citet{Kitchenham2004procedures}. We conducted a comprehensive literature search using Boolean search strategy in five databases, namely Web of Science, Scopus, ScienceDirect, PubMed, and Google Scholar (Table \ref{tab:boolean_search_strings}). Our search included all publications up to May 20th, 2023. 

Papers will be included if they meet the following criteria:
\begin{itemize}
    \item Focus on diagnostic tasks related to knee osteoarthritis (OA)
    \item Propose an end-to-end artificial intelligence (AI) model
    \item Utilize explainable AI (XAI) methods to provide explanations for the proposed model
    \item Not a review paper
    \item Published in English
\end{itemize}

\begin{figure}
    \centering
    \includegraphics[width=\linewidth]{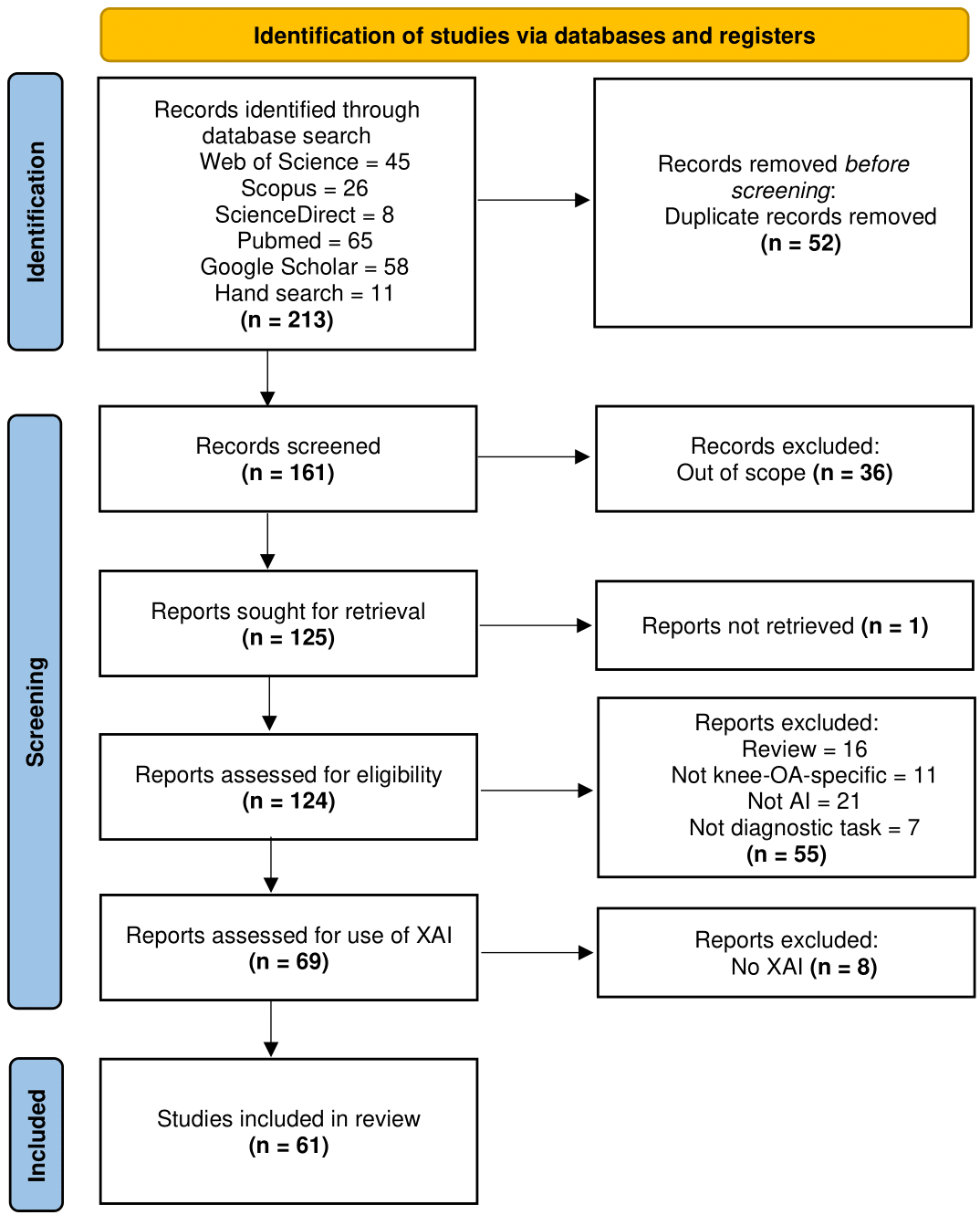}
    \caption{PRISMA flowchart depicting the study selection process for this systematic review.}
    \label{fig_prisma_flowchart}
\end{figure}

\begin{figure}
	\centering
            \includegraphics[width=\linewidth]{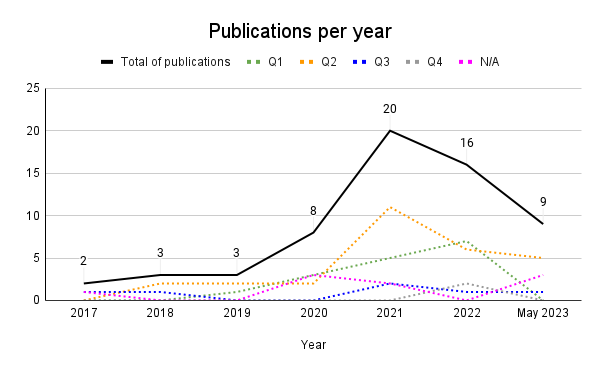}
	\caption{Trend of publications between 2017 to 2023.}
	\label{fig_trend_of_publication}
\end{figure}

Our review identified a total of 65 studies that presented at least one knee OA computer-assisted diagnostic system utilizing an end-to-end AI approach. Among these studies, 61 out of 69 (88.4\%) incorporated explainable AI (XAI) techniques and were included for our analysis (Figure \ref{fig_prisma_flowchart}). The earliest publication in this domain was found in 2017, which coincides with the introduction of popular XAI approaches such as gradient-weighted class activation map (GradCAM) \citep{Selvaraju2017} and self-attention mechanism \citep{Vaswani2017attention}. The introduction of these techniques sparked increased interest and discussion surrounding XAI in the field of knee OA diagnosis, therefore the publication trend experienced exponential growth from 2017 to 2021, as depicted in Figure \ref{fig_trend_of_publication}. Although there was a slight decrease in publications in 2022, the number of publications in high-quality (Q1) journals has been increasing steadily.

\begin{table}[width=.98\linewidth,cols=4,pos=h]
\caption{Boolean search strings employed for the corresponding bibliographic databases and search engines.}\label{tab:boolean_search_strings}
\begin{tabularx}{\tblwidth}{@{} lX@{} }
\toprule
\textbf{Database} & \textbf{Boolean search strings}\\
\midrule
Scopus & TITLE-ABS-KEY ( "knee"  OR  "tibiofemoral joint" )  AND  TITLE-ABS-KEY ( "osteoarthritis"  OR  "degenerative arthritis"  OR  "degenerative joint disease"  OR  "wear-and-tear arthritis" )  AND  TITLE-ABS-KEY ( "XAI"  OR  ( ( "explainable"  OR  "interpretable" )  AND  ( "AI"  OR  "artificial intelligen*"  OR  "deep learning"  OR  "machine learning" ) )  OR  "SHAP"  OR  "LIME"  OR  "gradcam"  OR  "grad cam"  OR  "heatmap"  OR  "saliency map"  OR  "attention map" ) \\
Web of Science & (((TS=(knee OR tibiofemoral joint)) AND TS=(osteoarthritis OR degenerative arthritis OR degenerative joint disease OR wear-and-tear arthritis)) AND TS=(XAI OR ((explainable OR interpretable) AND (AI OR artificial intelligen* OR deep learning OR machine learning)) OR SHAP OR LIME OR gradcam OR grad cam OR heatmap OR saliency map OR attention map)) \\
PubMed & (knee OR tibiofemoral joint) AND (osteoarthritis OR degenerative arthritis OR degenerative joint disease OR wear-and-tear arthritis) AND (XAI OR ((explainable OR interpretable) AND (AI OR artificial intelligen* OR deep learning OR machine learning)) OR SHAP OR LIME OR gradcam OR grad cam OR heatmap OR saliency map OR attention map) \\
ScienceDirect & (knee) AND (osteoarthritis) AND (XAI OR ((explainable OR interpretable) AND (AI OR artificial intelligen OR deep learning OR machine learning))) \\
Google Scholar & intitle:((knee) AND (osteoarthritis OR joint disease) AND (XAI OR ((explainable OR interpretable) AND (AI OR artificial intelligen OR deep learning OR machine learning)) OR SHAP OR LIME OR gradcam OR grad cam OR heatmap OR saliency map OR attention map)) \\
\bottomrule
\end{tabularx}
\end{table}

\section{General query on knee OA assessment}
\label{sec:general_query_on_knee_OA_assessment}
\begin{figure*}
\centering
\includegraphics[width=\textwidth]{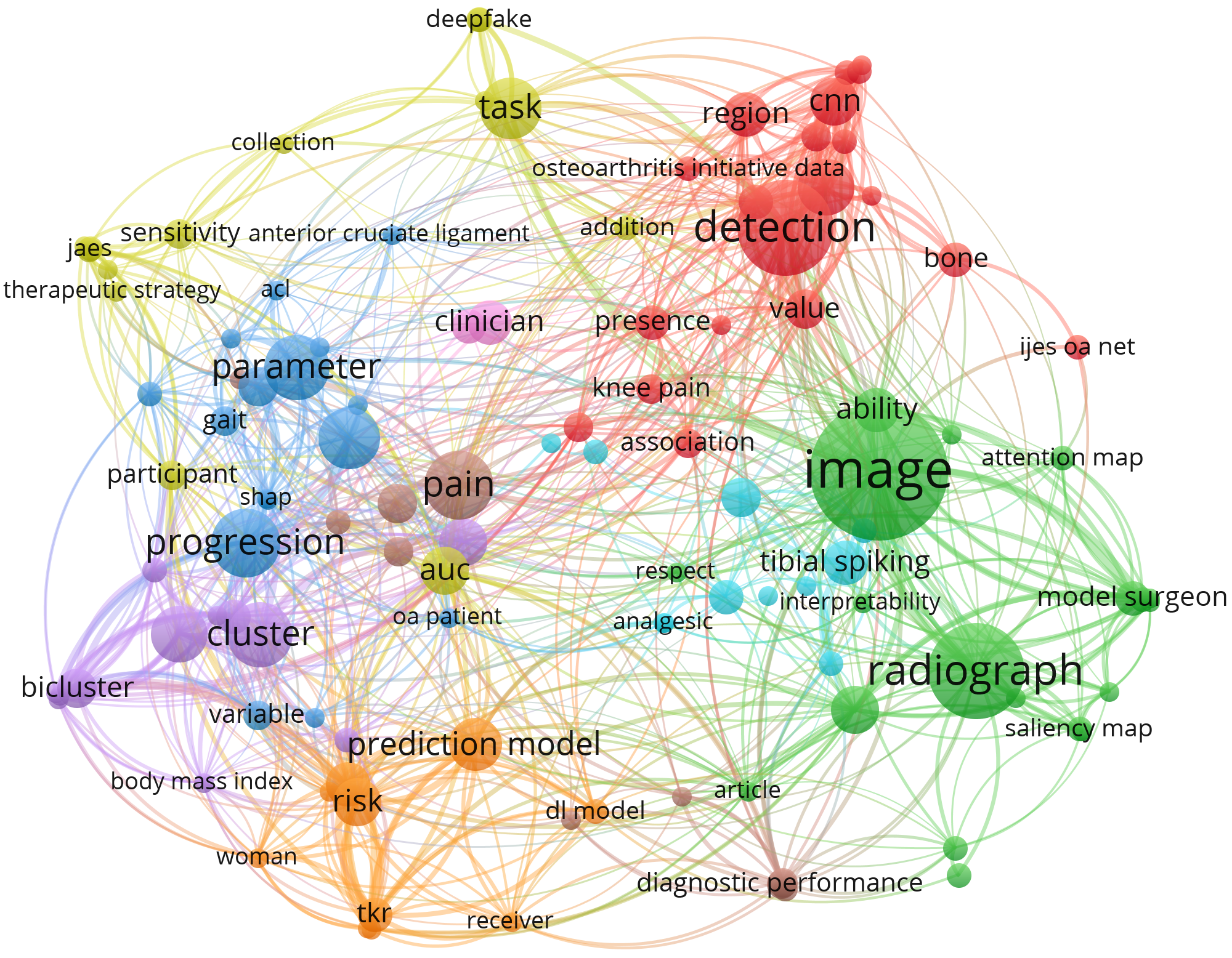}
\caption{Visual representation of the scientific landscape of the selected studies using VOSviewer's mapping function.}\label{fig_VOS_9_clusters}
\end{figure*}

We conducted an analysis of the general query to acquire an up-to-date comprehension of the topic on XAI application for knee OA diagnosis. This analysis aims to complement the qualitative literature review and provide valuable insights into the current state of research in this area. Co-occurrence analysis was performed using VOSviewer \citep{Van2010} to discover the relationships among terms extracted from the titles and abstracts of the selected studies. Out of the 1,455 terms identified, a subset of 190 terms with an occurrence frequency of at least three were chosen for analysis. Out of these 190 terms, we focused on the top 114 terms based on their relevance score, which fell within the top 60\% range. These 114 terms were then included in our analysis to gain insights into their co-occurrence patterns and relationships (Figure \ref{fig_VOS_9_clusters}). 
 
 As a result, the analysis revealed nine distinct clusters. Cluster 1 (20 items), Cluster 3 (16 items), Cluster 5 (12 items), and Cluster 7 (10 items) primarily focused on various aspects of knee OA symptoms and underlying risks, including bone and cartilage conditions, anterior cruciate ligament injury, demographics, and risk of OA deterioration, respectively. Cluster 2 (19 items) emphasized model interpretability and clinical practitioners. Cluster 4 (14 items), Cluster 8 (10 items), and Cluster 9 (2) was specifically associated with patient data. Cluster 6 (11 items) mainly encompassed studies related to automatic early diagnosis of knee OA. 
 
 The ten most cited terms included “image” (33 occurrences), “radiograph” (21 occurrences), “detection” (21 occurrences), “progression” (14 occurrences), "pain" (14 occurrences), “cluster” (13 occurrences), “parameter” (13 occurrences), “risk factor” (12 occurrences), "task" (12 occurrences), and “mri” (11 occurrences).

\section{Data for knee OA assessment}
\label{sec:data_for_knee_OA_assessment}
Knee OA is a complex and multifactorial disorder, and as such, a wide variety of data can be utilized to gain insights and explanations related to this health condition. In this review, we specifically focus on tabular data and image pixels. To track the evolving landscape of knee OA research, we performed an analysis of available datasets for knee OA assessment. This analysis provides a comprehensive understanding of the Western and Eastern data sources in knee OA diagnosis. We also highlight the role of datasets for generalizability and applicability of AI-based approaches.

\subsection{Tabular data}
Tabular data in knee OA research is a collection of structured information encompassing both objective and subjective measurements of the condition. Within this tabular data, we have identified six distinct domains: demographic, clinical, imaging, patient-reported outcomes, biomechanics, and biomarkers. Demographic data represents information about participants' characteristics, such as medical history, symptoms, demographics, nutrition, physical activity, comorbidity, and behavioral aspects. Clinical data involves physical exam and blood measures, outlining patients' essential health information. Imaging data consists of medial imaging outcomes and anthropometrics for quantifying anatomical structures. Patient-reported outcomes focus on data collected through questionnaires to assess patient-reported symptoms and health-related quality of life. Biomechanical data involves the mechanics and movement of the knee during various activities. Biomarkers data includes measurable indicators found in bodily fluids, offering insights into disease status and treatment response. A comprehensive comparison of the accessibility, cost, complexity, diagnostic error, adverse effects, risk of bias, and level of knowledge required for each domain is presented in Table \ref{tab:data_types}. This evaluation aids researchers in understanding the strengths and limitations of each data domain.

\subsection{Image pixels}
Image pixels in knee OA research consist of 2D or 3D data that allow for the visualization of human bone and tissue structures. This visual representation aids in gaining a deeper understanding of the anatomical aspects of the knee, enabling researchers to analyze and assess the condition more effectively. When paired with the tabular data of medical imaging outcomes, the combination of image pixels and structured data provides a more comprehensive approach to both qualitative and quantitative assessments of knee OA. This integration enhances the overall analysis and contributes to better knowledge discovery opportunities. However, handling image pixels data comes with challenges such as noise and resolution issues. High-resolution images offer improved visualization outcomes but also create a heavier computational load. Thus, a trade-off between image quality and computational efficiency needs to be carefully considered for practical implementation. 

\subsection{Geographical distribution and methodological insights}
\begin{figure}
\centering
\includegraphics[width=\linewidth]{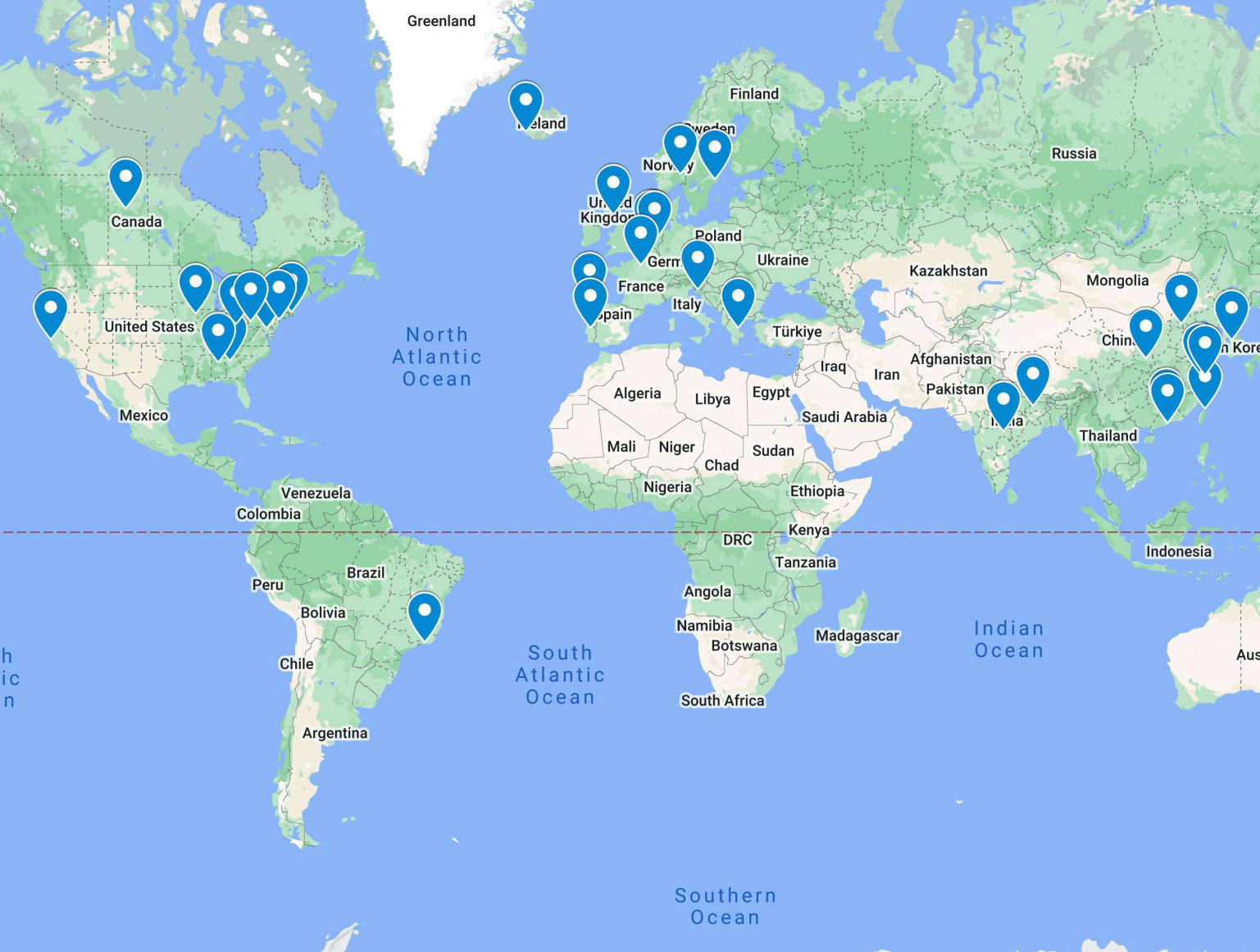}
\caption{Geographical distribution of OA data sources.}\label{fig_geographical_distribution_OA_data}
\end{figure}

Extensive data collection for OA research was conducted in both Western (n = 16) and Eastern (n = 10) countries (Fig. \ref{fig_geographical_distribution_OA_data}), with a particular focus on the United States and China regions. Existing research heavily relied on data from United States, where Caucassian is the largest population in the datasets. It is worth noting that limited research has been carried out in South American countries, and no research has been conducted in African countries. 

Overall, the identified datasets included a wide range of sample size, varying from 40 to 4,796 individuals. Notably, 62.3\% of the studies (38 out of 61) utilized the Osteoarthritis Initiative (OAI) dataset from United States for training or testing purposes. In 42 out of 61 studies, imaging data, primarily X-ray images (33 out of 42 studies), were utilized for clinical confirmation of OA disease. Approximately 38.7\% of the studies (24 out of 62) employed tabular or structured data, such as demographics, clinical characteristics, and laboratory examinations, to predict the risk of OA incidence. 

\begin{figure*}
\centering
\includegraphics[width=\linewidth]{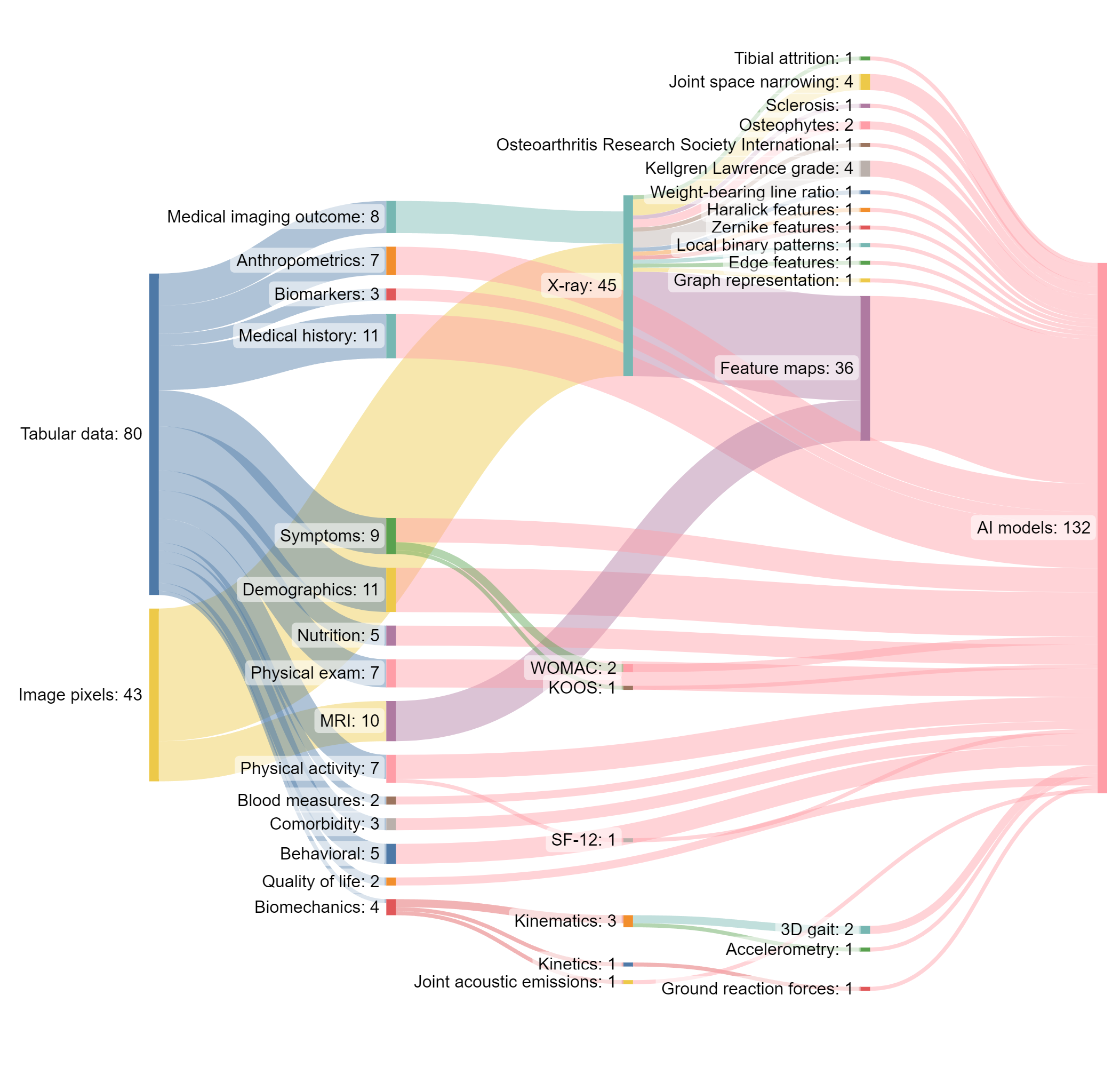}
\caption{Categorization and distribution of input data for AI models.}\label{fig5}
\end{figure*}

Utilizing data from a single channel, whether images or tables, poses significant challenges in knee OA research, as it may limit the comprehensive understanding of the condition. To address this limitation, two studies \citep{Tiulpin2019, Karim2021} adopted a data fusion approach, leading to the development of multimodal data models that maximize the utilization of patient information. By integrating diverse data types (Figure \ref{fig5}), these innovative approaches achieved more comprehensive and accurate predictions for knee OA assessment.

\subsection{Public datasets for knee OA assessment}
Due to ethical concerns and strict institutional regulations, there are limited public datasets available for knee OA assessment. Despite the availability of a greater number of private datasets, public datasets play a dominant role in establishing benchmark results and facilitating continuous improvement in the field of knee OA research. In this section, we present the publicly accessible datasets for knee OA diagnosis as outlined in
Table \ref{tab:Western_countries}.

\textbf{Osteoarthritis Initiative (OAI)} \citep{Osteoarthritis_Initiative} is an open access dataset provided by the National Institutes of Health (NIH). It focuses on identifying the most promising biomarkers of development and progression of symptomatic knee OA. This dataset includes 4,796 subjects between the ages of 45 and 79 years who either have knee OA or are at an increased risk of developing the condition. The data was collected from four clinical centers (Ohio State University, University of Maryland School of Medicine/Johns Hopkins University School of Medicine, University of Pittsburgh School of Medicine, and Brown University School of Medicine and Memorial Hospital of Rhode Island). Over a period of ten years, all participants underwent annual radiography and MRI scan of the knee, along with clinical assessments of disease activity. Furthermore, genetic and biochemical specimens were collected annually from all participants, providing rich data for researchers to explore novel knee OA diagnosis and treatment approaches.

\textbf{Multicenter Osteoarthritis Study (MOST)} \citep{Segalmulticenter} is a public dataset funded by the National Institutes of Health (NIH) and National Institute on Aging (NIA). The primary objective of this dataset is to study symptomatic knee OA in a community-based sample of adults with or at high risk of developing knee OA. About 3,026 subjects between the ages of 50 and 79 years from two clinical sites (Iowa City, Iowa and Birmingham, Alabama) participated the study. The dataset contains essential information related to biomechanical factors (such as physical activity-related factors), bone and joint structural factors (such as knee MRI assessment), and nutritional factors. 

\textbf{MRNet} \citep{Bien2018} is a collection of MRI data created by the Stanford University Medical Center. This dataset aims to investigate two common types of knee injuries: anterior cruciate ligament tears and meniscal tears which are contributing factors to knee OA disorder. The study involved 1,312 subjects and generated a total of 1,370 MRI scans. The MRI examinations were conducted using GE scanners (GE Discovery, GE Healthcare, Waukesha, WI) with a standard knee MRI coil and a routine non-contrast knee MRI protocol, comprising several key sequences: coronal T1 weighted, coronal T2 with fat saturation, sagittal proton density (PD) weighted, sagittal T2 with fat saturation, and axial PD weighted with fat saturation. Among the knee examinations, about 56.6\% were performed using a 3.0 Tesla magnetic field, while the remaining used a 1.5 Tesla magnetic field. Furthermore, the authors provided a benchmark MRNet single model, intended to support further research endeavors in the field.

\textbf{FastMRI+} \citep{Zhao2022fastmri+} is a publicly available MRI dataset that extended the work of the FastMRI dataset \citep{Knoll2020fastmri}. This extended dataset includes 1,172 MRI scans acquired at 1.5 or 3.0 Tesla and provides 22 different pathology labels in knee anatomical areas such as bone, cartilage, ligament, meniscus, and joint. Notably, many of the pathologies, such as cartilage loss and joint effusion, are closely related to knee OA. Each knee MRI scan comprises a single series of coronal images in PD or T2-weighted sequence. The primary focus of the FastMRI+ dataset is to facilitate the study of MRI image reconstruction, particularly in regions that could potentially contain clinical pathology. This dataset provides detailed pathology labels, researchers can explore and develop advanced image reconstruction techniques that cater to specific clinical conditions.

\textbf{Cohort Hip and Cohort Knee (CHECK)} \citep{Wesseling2016, Wang2022machine} is a research initiative sponsored by the Dutch Arthritis Foundation, in collaboration with ten general and university hospitals in The Netherlands, situated in semi-urbanized regions. The study recruited a total of 1,002 subjects aged between 45 and 65 years. The primary goal of this dataset is to explore and analyze the clinical, biochemical, and radiographic signs and symptoms associated with early OA. Moreover, the dataset aims to identify prognostic factors that may contribute to the diagnosis and progression of OA. The study spans a duration of seven years, during which 846 subjects actively participated in annual clinic visits, providing valuable longitudinal data for comprehensive OA research.

\textbf{Private research at Danderyd University Hospital} is used in \citet{Olsson2021} to develop a predictive model for classification of OA stage. The dataset consists of 6,103 X-ray images acquired from Danderyd University Hospital. Unlike other datasets that undergo extensive preprocessing for artifact removal, this dataset used the entire image series, including X-ray images with visual disturbances like implants, casts, and non-degenerative pathologies. This unique approach provides a more realistic representation of clinical scenarios and enhances the dataset's value for studying OA progression and prediction in real-world conditions.

\textbf{Mendeley VI} is a unique public dataset that focuses on the Eastern population. It contains 1,650 X-ray images collected from Indian institutions. The X-ray images were captured using the PROTEC PRS 500E X-ray machine. All images are 8-bit grayscale and have been cropped to focus on the cartilage region. They have been manually annotated by two experienced medical experts with their respective Kellgren and Lawrence grades. The intention of this dataset is to facilitate in the development of AI models for classifying osteoarthritis severity.

\begin{table*}[]
\caption{Comparison of input data types in knee OA diagnosis}
\label{tab:data_types}
\resizebox{\textwidth}{!}{%
\begin{tabular}{cccccccc}
\hline
\textbf{Input data type}          & \textbf{Accessibility} & \textbf{Cost} & \textbf{Complexity} & \textbf{Diagnostic error} & \textbf{Adverse effects} & \textbf{Bias} & \textbf{Knowledge required} \\ \hline
Demographic            & High                   & Low            & Low                 & Low                       & Low                      & Low            & Low                         \\ 
Clinical               & High                   & Moderate       & Moderate            & Low                       & Low                      & Low            & Moderate                    \\ 
Imaging                & Moderate               & High           & High                & Moderate                  & Low                      & Moderate       & High                        \\ 
Patient-Reported Outcomes   & Moderate               & Low            & Low                 & Moderate                  & Low                      & Low            & Low                         \\ 
Biomechanics      & Low                    & Low            & Moderate            & Low                       & Low                      & Low            & Low                         \\ 
Biomarkers            & Low                    & High           & High                & Moderate                  & Low                      & High           & High                        \\ \hline
\end{tabular}
}
\end{table*}

\begin{table*}[]
\caption{List of OA-related data sources and descriptions in Western countries.}
\label{tab:Western_countries}
\resizebox{\textwidth}{!}{%
\begin{tabularx}{\textwidth}{p{5.5cm} p{1.5cm} p{2.0cm} p{4.0cm} p{2.0cm} }
\hline
\textbf{Data source (website)} & \textbf{Year of data} & \textbf{No. of subjects (age range)} & \textbf{Data types} & \textbf{Geographic representation} \\ \hline
Osteoarthritis Initiative (OAI) \citep{Osteoarthritis_Initiative} (\url{https://nda.nih.gov/oai/}) & 2004-2015 & 4,796 (45-79) & Image data, tabular data at baseline, and 12, 24, 36, 48, 60, 72, 84, 96, and 108 months & USA \\ \hline
Multicenter Osteoarthritis Study (MOST) \citep{Segalmulticenter} (\url{https://most.ucsf.edu/}) & 2003-2018 & 3,026 (50-79) & Image data, tabular data at baseline, 15, 30, 60, 72, and 84 months & USA \\ \hline
MRNet \citep{Bien2018} (\url{https://stanfordmlgroup.github.io/competitions/mrnet/}) & 2001-2012 & 1,312 (-) & Image data: 1,370 MRI images & USA \\ \hline
FastMRI+ \citep{Zhao2022fastmri+} (\url{https://github.com/microsoft/fastmri-plus} and \url{https://fastmri.med.nyu.edu/}) & N/A & - (-) & Image data: 1,172 coronal MRI scans either proton density-weighted and T2-weighted & USA \\ \hline
Cohort Hip and Cohort Knee (CHECK) \citep{Wesseling2016, Wang2022machine} & 2002-2012 & 1,002 (45-65) & Image data and tabular data at baseline, 2, 5, 8, and 10 years & The Netherlands \\ \hline
Private research at Danderyd University Hospital \citep{Olsson2021} (\url{https://datahub.aida.scilifelab.se/10.23698/aida/koa2021}) & 2002-2016 & - (-) & Image data: 6,403 X-ray images & Sweden \\ \hline
Mendeley VI \citep{Gornale2020digital} (\url{https://data.mendeley.com/datasets/t9ndx37v5h/1}) & N/A & - (-) & Image data: 1,650 X-ray images & India \\ \hline
\end{tabularx}
}
\end{table*}

\section{Classification systems for knee OA conditions}
\label{sec:classification_systems_for_final prediction_of_knee_OA_conditions}
In this section, we conducted a comparison of the employment of classification systems from the medical domain that establish the ground truth data for predictive models. Approximately half of the studies (32 out of 61) utilized medical experts' knowledge for classification or clustering tasks. Within this subset of studies, 28 employed Kellgren Lawrence (KL) grading system to rate the OA severity. Original Kellgren Lawrence (KL) grading system comprises five ordinal classes based on composite score of radiographic OA symptoms. However, the number of classes used in the top layer of the KL prediction models varied from two to five across the reviewed studies, depending on their respective research purposes. A commonly used standard threshold for radiological OA is a KL$\geq$2. Most of the studies (18 out of 28) were dedicated to developing AI models specifically for the five-grade KL classification. Binary classification was designed to identify the presence of OA (KL1 to KL4) (4 out of 28) or early OA (KL2) (4 out of 28). In addition, one study classified the change in KL grade after 60 months. 

Besides KL grading, there was one study employed Osteoarthritis Research Society International (OARSI) atlas joint-space narrowing for medial tibiofemoral OA. Another research developed radiographic spiking criteria to guide the generation of ground truth data \citep{Patron2022automatic}. Whole-Organ Magnetic Resonance Imaging Score (WORMS) (n = 1) and MRI Osteoarthritis Knee Score (MOAKS) (n = 1) were employed for knee OA detection on MRI data. Both classification systems emphasized cartilage damage related to OA. 

In contrast, patient-reported outcome measure were only used in four studies. Western Ontario and McMaster Universities Osteoarthritis Index (WOMAC) (n = 2) and the Knee Injury and Osteoarthritis Outcome Score (KOOS) (n = 1) were frequently employed as patient-reported outcome measures, particularly for assessing pain. However, due to their subjective nature, the analysis process for these measures was complex and required careful statistical analysis \citep{Pierson2021}. Moreover, transforming the data from these measures into a format suitable for modeling presented a significant challenge. \citet{Morales2021} established a direct binary classification system for chronic knee pain based on patient self-reporting. They defined chronic knee pain as pain that persists for more than half of the days in a month for at least six out of the past 12 months. Moreover, two studies utilized knowledge-based and patient-based outcome measures \citep{Zeng2022, Chan2021}.

\section{XAI approaches for knee OA assessment}
\label{sec:XAI_approaches}
The role of XAI in knee OA assessment is to offer comprehensible explanations regarding the input data. These explanations are intended to be understood by humans. Thus, we take into account the interests of data scientists and domain experts in the development of XAI methods.

For data scientists, knowing the internal workings of the model and comprehending how the data is applied are crucial for improving the model's performance and preventing overfitting. This knowledge enables them to fine-tune the model, optimize its architecture, and make informed decisions during the development process. Post-hoc explanations may be of lesser concern to them, as they prioritize optimizing the model itself. 

On the other hand, domain experts especially medical experts who may not have the technical expertise of data scientists are more interested in understanding how and why a model generated a particular result. They seek clear and interpretable explanations to trust the model's decisions and insights. Knowing the key characteristics that led to a conclusion helps them validate the model's outputs and make informed decisions based on the AI system's recommendations.

\begin{figure*}
\centering
\includegraphics[width=\textwidth]{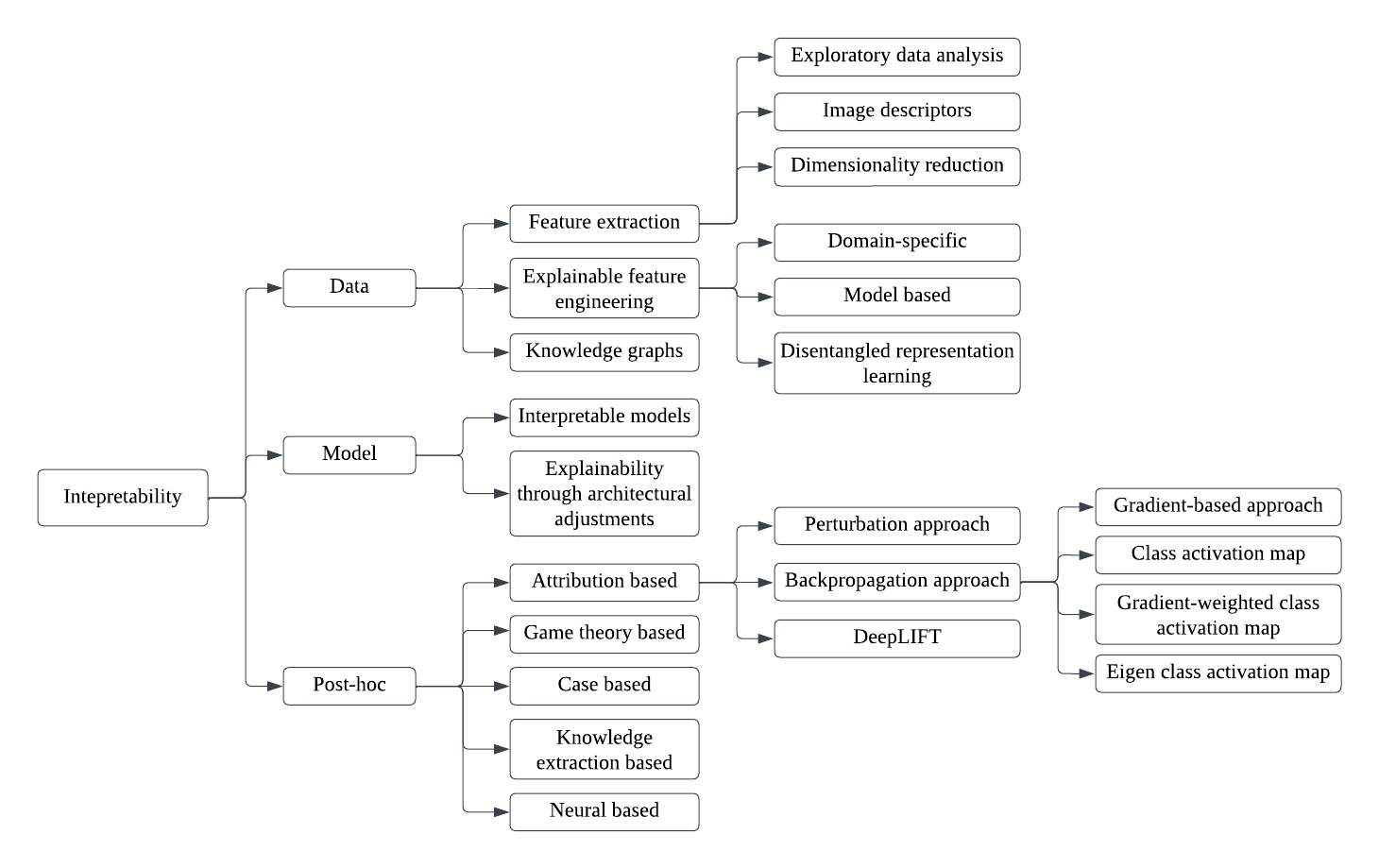}
\caption{Proposed XAI taxonomy.}\label{fig_XAI_taxonomy}
\end{figure*}

By considering the specific needs and interests of both data scientists and domain experts, we propose the XAI taxonomy as shown in Figure \ref{fig_XAI_taxonomy} to provide valuable insights into the diverse requirements of different stakeholders. Understanding data interpretability, model interpretability, and post-hoc interpretability, along with XAI evaluation approaches, is crucial in building transparent, trustworthy, and effective AI models that cater to various real-world applications.

\subsection{Data interpretability}
The importance of data interpretability arises from the substantial impact of the training dataset on an AI model's behavior. To facilitate a better understanding of the input data, numerous data analysis techniques and mathematical algorithms have been developed to quantify the intrinsic data characteristics. In the context of knee OA, data interpretability can uncover valuable clinical patterns that might not have been captured in traditional evidence-based research. This can empower the researchers to glean new insights and knowledge from the data, contributing to more informed and effective decision-making in knee OA assessment.

In the following sections, we will discuss a few approaches that provide interpretability for knee OA data. This includes feature extraction, explainable feature engineering, and knowledge graphs, which are widely recognized as pre-modelling approaches. These approaches help extract useful information from the data and represent different steps in achieving data interpretability. Feature extraction extracts relevant features, explainable feature engineering transforms data for better understanding, and knowledge graphs connect related points for a comprehensive disease overview.

\subsubsection{Feature extraction}
Feature extraction plays a critical role in capturing a representative set of features. In our survey, we found three types of feature extraction: exploratory data analysis, image descriptors, and dimensionality reduction. 

\paragraph{\textbf{Exploratory data analysis}}
Exploratory data analysis (EDA) is a data analysis approach that involves summarizing the main characteristics of the data and visualizing the data summary using appropriate representations \citep{Sahoo2019exploratory}. EDA is an essential process for understanding the structure and distribution of tabular data, as well as identifying important features and patterns that can guide subsequent analysis. The significant contribution of EDA in knee OA data is analysis of population-based samples to provide the disease overview and to detect biases in data \citet{Angelini2022}.

General EDA outcomes included dimensions, mean, median \citep{Angelini2022}, standard deviation, range, and missing samples. To deal with missing samples, \citet{Angelini2022} implemented imputation models using random forest (RF) and k-nearest neighbor (KNN) regression models, which were further refined through a bootstrapping-like procedure. \citet{Jakaite2021deep} performed an analysis on the brightness value distributions in the lateral and medial sides of the OA and control groups. The results showed that the mean brightness in the OA group was higher than in the control group in both sides. The observation suggested that the higher mean brightness may be indicative of increased bone density in patients from the OA case group.

\paragraph{\textbf{Image descriptors}}
Image descriptors are typically used to capture and describe the shape of an object in an image. \citet{Jakaite2021deep} utilized Zernike moments (Equation \ref{zernike_moments}) to capture knee X-ray textural details at the bone microstructural level. By using this image descriptor and the Group Method of Data Handling (GMDH), they were able to effectively identify patients at risk of early knee OA, even with a relatively small dataset of 40 samples.

\begin{subequations}
\label{zernike_moments}
The Zernike moments ${\displaystyle \ A_{nm}}$ are defined as:
\begin{equation}
    A_{nm}=\frac{n+1}{\pi }\sum _x{}\sum _y{f(x,y)V^*_{nm}(\rho ,\theta )}
\end{equation}
\noindent
where ${\displaystyle \ f(x, y)}$ represents the image, ${\displaystyle \ {n}}$ denotes the number of order, ${\displaystyle \ m}$ denotes the number of repetition, ${\displaystyle \ V_{nm}}$ represents orthogonal complex polynomials, ${\displaystyle \ \rho}$ represents the length of a vector to a ${\displaystyle \ (x, y)}$ pixel, and ${\displaystyle \ \theta}$ represents the angle between x-axis and ${\displaystyle \ \rho}$. \\

The orthogonal complex polynomials ${\displaystyle \ V_{nm}}$ are defined as:
\begin{equation}
    V_{nm}(x,y)=V_{nm}(\rho ,\theta )=R_{nm}(\rho )e^{(-jm\theta )}
\end{equation}
\noindent
where ${\displaystyle \ R_{nm}}$ represents radial polynomials, ${\displaystyle \ \rho}$ represents the length of a vector to a ${\displaystyle \ (x, y)}$ pixel, and ${\displaystyle \ \theta}$ represents the angle between x-axis and ${\displaystyle \ \rho}$. \\

The radial polynomials ${\displaystyle \ R_{nm}}$ are defined as:
\begin{equation}
    R_{nm}=\sum _{k=0}^{(n-|m|)/2}\frac{(-1)^k(n-k)!}{k!((n+|m|)/2-k)!((n-|m|)/2-k)!}\rho ^{n-2k}
\end{equation}
\noindent
where ${\displaystyle \ {n}}$ denotes the number of order and ${\displaystyle \ m}$ denotes the number of repetition. \\

\end{subequations}

A more detailed bone analysis work was performed by \citep{Bayramoglu2020}. The authors conducted a comparison of five image descriptors: local binary pattern (LBP), fractal dimension (FD), Haralick features, Shannon entropy, and histogram of gradient (HOG). Based on their findings, they recommended the use of LBP as it preserved the most discriminative features among the descriptors. They also pointed out that LBP and HOG descriptors are less sensitive to changes in radiographic acquisition protocols and could be applied in clinical decision support tools in the future.

Besides texture analysis, image descriptor could be used to extract object edge information. Adaptive Canny algorithm was employed to extract the edges of the knee joint from X-ray images by dynamically adjusting the threshold values based on the local image characteristics \citep{Farajzadeh2023ijes}. The low ${\displaystyle \ \alpha}$ and high adaptive thresholds ${\displaystyle \ \beta}$ are defined as:
\begin{subequations}
\label{adaptive_canny}
\begin{align}
    \alpha &= \max\left(0, (1-\sigma) \times \text{median}(x_i)\right) \\
    \beta &= \min\left(255, (1+\sigma) \times \text{median}(x_i)\right)
\end{align}
\end{subequations}
\noindent
where ${\displaystyle \ \alpha}$ denotes upper limit pixel value, ${\displaystyle \ \beta}$ denotes bottom limit pixel value, and ${\displaystyle \ x_i}$ represents median pixel value.

\paragraph{\textbf{Dimensionality reduction}}
OA datasets are typically complex and multidimensional, containing a vast amount of variables. Visualizing such high-dimensional data can be challenging since human perception is limited to three dimensions. Hence, researchers tend to find lower-dimensional representations of the original data \citep{Murdoch2019definitions}. Dimensionality reduction techniques are employed to reduce the number of parameters while preserving the underlying structure as much as possible. Two commonly used methods in this field are Principal Component Analysis (PCA) \citep{Angelini2022} and t-Distributed Stochastic Neighbor Embedding (t-SNE) \citep{Chen2019fully,Chan2021,Li2023deep,Wang2023confident,Wang2023transformer}. 

\subsubsection{Explainable feature engineering}
There are two main approaches developed for explainable feature engineering: domain-specific methods and model-based methods. Another emerging approach in explainable feature engineering is disentangled representation learning, which has gained traction with the introduction of various generative models.

\paragraph{\textbf{Domain-specific}}
Domain-specific approaches for knee OA diagnostic task utilize the knowledge and expertise of medical experts, along with insights derived from EDA to extract features. Many studies in this field have focused on developing knee-specific approaches that capture and characterize key aspects on bone and cartilage, as well as the limb alignment. \citet{Du2018novel} developed a measure called cartilage damage index (CDI) to quantify cartilage thickness by measuring specific informative locations on the reconstructed cartilage layer instead of evaluating the entire cartilage. In a cartilage assessment conducted by \citet{Ciliberti2022}, two volumetric analyses were employed. The first analysis focused on wall thickness, where the cartilage mesh was examined, and the thickness of each element was calculated from surface to surface. The hypothesis underlying this analysis was that patients with degenerative and traumatic cartilages would exhibit thinner cartilage in specific regions compared to the control group. The second analysis focused on cartilage curvature by measuring the Gaussian curvature of cartilage element based on its neighboring elements. This analysis hypothesized that areas with higher cartilage degradation would exhibit increased curvature due to the formation of holes and depressions surrounding those regions. In \citep{Morales2021}, cartilage thickness was determined for femoral, tibial, and patellar cartilage masks per sagittal slice by performing an Euclidean distance transform along the morphological skeleton of each mask. Furthermore, the shape of the bone was characterized by measuring the distance from the bone surface of each bone mask to its volumetric centroid. A recent study by \citep{Zhuang2022} proposed a unified graph representation approach to construct personalized knee cartilages that are attached to the femur, tibia, and patella, respectively. They used the patient-specific cartilage graph representation to guide their DL model. Additionally, to assess the coronal limb alignment through radiographic means, weight-bearing line (WBL) ratio was derived by calculating the ratio between the crossing point of the mechanical axis, measured from the medial edge of the tibial plateau, and the total width of the tibial plateau \citep{Moon2021}.

\paragraph{\textbf{Model based}}
 Model-based feature engineering leverages an automatic approach to unveil the inherent structure of a dataset, leading to the extraction of relevant and informative features \citep{Murdoch2019definitions}. One such example is unsupervised clustering, a technique that groups similar data points together based on their intrinsic characteristics, without the need for labelled target variables \citep{Murdoch2019definitions}. For instance, \citet{Morales2021} developed a fully automatic landmark-matching algorithm based on Coherent Point Drift to map the bone surfaces into reference space. \citet{Bayramoglu2020} used simple linear iterative clustering based superpixel segmentation to extract the region of interest as a pre-processing strategy. \citet{Angelini2022} applied k-means clustering to analyze biochemical marker data and figure out prominent subgroups among patients with OA. This approach enabled them to identify three dominant OA phenotypes. \citet{Nelson2022biclustering} conducted similar work using biclustering, but their work was extended to more inclusive clinical data, including demographics, medical history, symptoms, physical activity, physical exam, and medical imaging outcome. Through their analysis, they identified two significant clusters. One cluster represented individuals who exhibited structural progression over time but experienced improvements in pain. The other cluster represented individuals who had stable pain scores and were less affected by OA. Additionally, model-based feature engineering techniques were employed to analyze gait data, which is known for its complexity with multidimensional and time-series properties. \citet{Leporace2021} utilized self-organizing maps (SOM) on principal components to detect gait similarity patterns in individuals with high grade OA. The resulting patterns were visualized using a unified distances matrix (U-matrix). Subsequently, the U-matrix was subjected to the k-means clustering algorithm, leading to the formation of four distinct gait kinematic clusters.

\paragraph{\textbf{Disentangled representation learning}}
Disentangled representation learning is a significant and closely related area of research that focuses on acquiring a dataset representation where the generative latent variables are disentangled or separated. Latent variables in this context can be regarded as interpretable or explainable features of the dataset. \citet{Prezja2022deepfake} were pioneers in applying the DeepFake concept in this medical domain, specifically by utilizing Wasserstein generative adversarial neural networks with gradient penalty (WGAN-GP). Their model managed to preserve important OA anatomical information during the generation process. The authors utilized DeepFake generated data to substitute real data during the training of a pre-trained VGG model for classification task. Remarkably, they achieved a mere 3.79\% decrease in accuracy compared to the baseline when classifying real OA X-rays. \citet{Wang2023key} advanced the field of disentangled representation learning by introducing a novel approach called key-exchange convolutional autoencoder (KE-CAE). This method was designed to extract specific radiograph features related to early knee osteoarthritis (OA) from latent space through cross image reconstruction. Their proposed approach successfully captured crucial information from radiographs that represents early knee OA, enabling effective analysis. Notably, their model not only achieved high-quality reconstruction of the original images but also generated synthetic images that accurately represented different stages of knee OA. This noteworthy contribution holds promise for the early detection and diagnosis of knee OA.

\subsubsection{Knowledge graphs}
\begin{figure*}
\centering
\includegraphics[width=\linewidth]{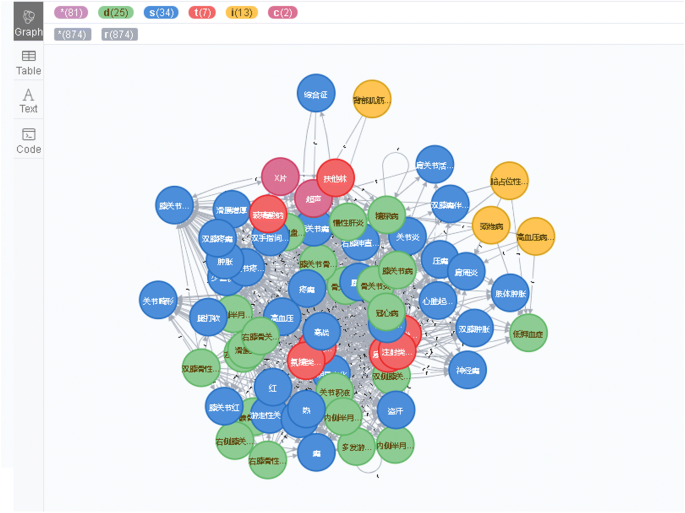}
\caption{Example of knowledge graph for knee OA. Each circle represents a specific Chinese term associated with knee OA. Adapted from \citep{Li2020}}\label{fig_knowledge_graph}
\end{figure*}

Knowledge graph (Figure \ref{fig_knowledge_graph}) is a structured representation of knowledge that captures relationships between entities in a particular domain. \citet{Li2020} established a medical knowledge graph using unstructured data from an electronic medical record (EMR) database. The EMR data was in Mandarin language, which posed a computational challenge for processing Mandarin words. To address this, the authors adopted five feature extraction methods, including bi-directional long short-term memory (Bi-LSTM), bag of characters, natural language processing with the Chinese Academy of Sciences Word Segmentation Tool, dictionary features, and the k-means algorithm for word clustering. These diverse feature sets were utilized in the conditional random field (CRF++) algorithm for entity recognition. Following entity recognition, the authors combined the extracted features and implemented a CNN model, comprising a convolutional layer, pooling layer, fully connected layer, and softmax classifier, to extract entity relations from the identified entities. This step allowed for a deeper understanding of the interconnections within the medical data. In the final stages of building the medical knowledge graph, the authors employed Neo4j graph database. They achieved this by batch importing the previously identified medical entities and their corresponding relationships into the Neo4j database, forming a comprehensive and interconnected representation of medical knowledge in knee OA domain. The resulting knowledge graph encompassed 2,518 distinct entities and an impressive 29,972 different relationships related to knee OA condition. The knowledge graph spans a diverse range of entity types, comprising 368 diseases, 706 symptoms, 421 treatments, 859 examination descriptions, 72 examinations, 43 aggravating factors, 35 mitigating factors, and 14 inducing factors. This comprehensive repository of information serves as a valuable resource, empowering researchers and medical practitioners to gain deeper insights into knee OA.

\subsection{Model interpretability}
While clean and carefully prepared data, aided by data interpretability techniques, is crucial for training models, it is equally important for the model itself to possess a clear understanding. Without this understanding, developers may face challenges when incorporating their domain knowledge into the learning process to achieve improved results. Therefore, alongside data interpretability, model interpretability plays a vital role. 

In many instances, analyzing outputs or examining individual inputs is insufficient for comprehending why a training procedure failed to yield the desired outcomes. In such cases, it becomes necessary to investigate the training procedure in the model. The objective of model explainability is to develop models that are inherently more interpretable and understandable. This approach is also called intrinsic XAI.

\subsubsection{Interpretable models}
Interpretable models, also known as white-box models, are models that provide self-explanatory insights \citep{Du2019techniques}. Examples of such models include rule-based model, linear regression, logistic regression, and decision trees. 

In the realm of rule-based models, \citet{Pierson2021} devised an objective algorithm for pain prediction and compared it to a general KL grade-based algorithm. Their proposed algorithm incorporated racial disparities (Black versus non-Black) and two socioeconomic measures, namely annual income below \$50,000 and educational attainment (college graduation). The authors examined the differences in pain scores between groups and quantified the pain disparities using non-parametric means. By employing a regression model, the proposed algorithm successfully addressed the inequalities faced by under-served patients. 

\citet{Zeng2022} utilized binary logistic regression to detect knee OA and recommend appropriate treatment options, including conservative or surgical approaches. Although the authors claimed the interpretability of their model, however they did not provide detailed analysis or explanations to support their claim.

\begin{figure*}
\centering
\includegraphics[width=\linewidth]{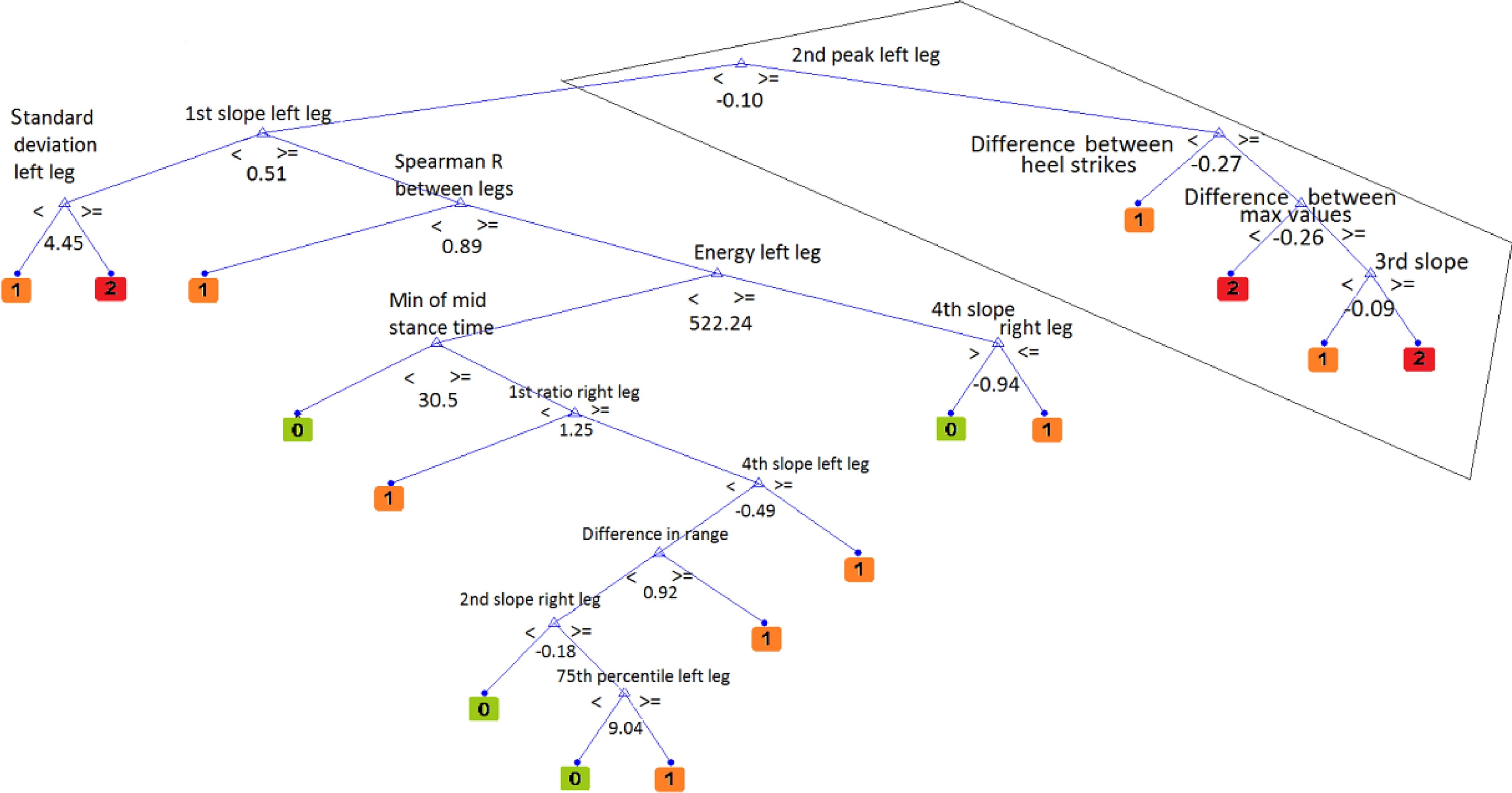}
\caption{Example of regression tree. Adapted from \citep{Kotti2017}}\label{fig_regression_tree}
\end{figure*}

In terms of tree-based models, \citet{Kotti2017} utilized a regression tree to analyze and interpret the rule induction process for detecting OA cases from a biomechanical perspective. A random subset of parameters extracted from ground reaction forces in the z-axis was employed to construct the regression tree, as illustrated in Figure \ref{fig_regression_tree}. This approach provided insights into the biomechanical factors that may contribute to the presence of OA and offered a means of interpreting the rule induction process in the context of OA detection. 

\citet{Liu2018interpretable} employed splitting nodes algorithm to assess the importance of each feature in a tree generated by eXtreme Gradient Boosting (XGBoost). They found that demographics and anthropometric factors had a significant influence on determining OA status, but acknowledged that these factors are not exclusive to OA and contribute to various clinical issues like pain and disability. Instead, the authors emphasized three other categories: comorbidity, blood measures, and physical activity measures. These categories were closely linked to the risk of experiencing side effects from analgesics in OA patients.

Despite the use of white-box mechanisms, relying solely on interpretable models may not provide sufficient explanation for complex models, particularly in scenarios with high-dimensional and heterogeneous data. To address this limitation, the application of regularization techniques becomes necessary during model training. Regularization helps control the number of relevant input features by introducing penalties or constraints, ensuring that the model focuses on the most important variables. For example, \citet{Kokkotis2020machine} used a robust methodology to process 707 features from multidisciplinary settings. They employed six feature selection techniques, including filter algorithms, wrapper approach, and embedded techniques, and ranked features based on a majority vote scheme. This process identified 40 relevant risk factors, resulting in a classification accuracy of 77.88\% using logistic regression. 

However, one limitation of the majority voting approach is that it treats all models in the ensemble equally, without considering the possibility of weak predictions. To address this limitation, \citet{Ntakolia2021identification} introduced a Fuzzy ensemble approach to optimize the model and improve decision-making by considering the reliability and uncertainty of individual predictions. Additionally, Lu et al. (2022) demonstrated the effectiveness of recursive feature elimination (RFE), which considers the intrinsic characteristics of the data and model to select an optimal feature combination. RFE iteratively eliminates less relevant features, resulting in an informative subset that contributes significantly to the model's performance.  

\subsubsection{Explainability through architectural adjustments}
Attention mechanisms could introduce certain level of explainability and have revolutionized the utilization of DL algorithms \citep{Zhang2020, Schiratti2021, Wang2021automatic, Feng2021automated, Zhuang2022, Huo2022automatic}. \citet{Zhang2020} utilized the convolutional block attention module (CBAM) to implement an attention mechanism. Their CBAM consisted of both channel and spatial attention modules. By incorporating the module into ResNet34, the proposed approach identified the most relevant channel and spatial parts that contributed significantly to the final prediction and helped to enhance the model's performance by focusing on the most informative features in the input data during the training process.

In the study conducted by \citet{Feng2021automated}, the channel attention module within the CBAM was enhanced by incorporating additional non-linear layers after fusing the channel weights from dual branches. This modification increased the expressiveness of the CBAM network and improved the model's accuracy in detecting potential lesions in knee X-ray images. Similarly, \citet{Schiratti2021} incorporated a gated attention mechanism to calculate attention scores for individual image slices, which can be interpreted as indicators of their importance. These scores were then utilized in the classification sub-model. Self-attention mechanism was implemented by \citet{Wang2021automatic} by integrating a visual transformer after their deep learning model. Their approach effectively captured the interrelationship among imaging features from multiple regions. 

Alternatively, \citet{Zhuang2022} proposed a self-attention-based network, namely CSNet that has been designed in a layer-by-layer manner. Each layer incorporated patch convolution to extract local appearance features from individual vertices and graph convolution to facilitate communication among the vertices. The self-attention mechanism was employed in each layer to enhance the model's ability to capture information from the cartilage graph. The final assessment of knee cartilage defects was obtained by pooling information from all vertices in the graph, and the CSNet also allowed for easy 3D visualization of the defects, showcasing its interpretability. In contrast to previous approaches that did not take semantic information into account, a recent study by \citet{Huo2022automatic} introduced the use of an online class activation mapping (CAM) module to specifically direct the network's attention towards the cartilage regions.

\subsection{Post-hoc interpretability}
Post-hoc XAI methods were found to be more commonly employed than intrinsic XAI methods in those studies. These post-hoc methods provide an external explanation of the AI model's decisions after it has made predictions. It involves querying the trained model and constructing a white-box surrogate model to extract the underlying relationships the model has learned \citep{Murdoch2019definitions}. These methods could gain insights into the model's decision-making process by analyzing its predictions on specific instances, without altering the original model architecture. In contrast, intrinsic XAI focuses on designing AI models with inherent interpretability right from the model's architecture and design \citep{Du2019techniques}. These models are built with specific structures or components that naturally provide transparency and understandability in their decision-making process.

\subsubsection{Attribution based}
\paragraph{\textbf{Perturbation approach}}
Perturbation is a simple and effective method for computing the impact of changing input features on the output of an AI model \citep{Giuste2023}. It involves manipulating certain input features, running the forward pass, and measuring the difference from the original output. The importance of the input features can be ranked based on their effect on the output. \cite{Pierson2021} demonstrated a region-wise method to visualize image areas that influenced predictions made by a neural network. To do this, the image regions are "masked" out by replacing them with a circle, and the value of the circle was set to the mean pixel value for the image. Gaussian smoothing was applied to prevent sharp boundaries. The neural network's predicted pain score was then compared between the masked image and the original image, and the absolute change in the predicted pain level was computed. This process was repeated for a 32x32 grid of regions that evenly tiled the 1024x1024-pixel image. This has allowed a heatmap analysis that revealed how much masking each region of the image affected the neural network's prediction. 

\paragraph{\textbf{Backpropagation approach}}
Backpropagation approach can be further divided into gradient-based approach, class-activation map, and gradient-weighted class activation map. Nearly half of the studies (30 out of 61) employed backpropagation XAI approach to explain the predictions of AI models as described in Table \ref{tab_attribution_based_XAI}.

\noindent
\textbf{Gradient-based approach} focus solely on the gradient information when assessing the impact of modifying a specific pixel on the final prediction.  Integrated Gradients is a specific technique within this approach. Another technique, namely SmoothGrad was introduced with the intention of reducing visual noise \citep{Smilkov2017smoothgrad} and used by \citet{Tack2021} for MRI study.

\noindent
\textbf{Class activation map} (CAM) utilize global average pooling to compute the spatial average of feature maps in the final convolutional layer of a CNN \citep{Zhou2016learning}. Three studies \citet{Chang2020,Huo2022automatic,Dunnhofer2022deep} that focused on MRI data used CAM approach to analyze the prediction outcomes.

\noindent
\textbf{Gradient-weighted class activation map} (GradCAM) is an extension of CAM, which is a technique that does not depend on a specific architecture. The principle of GradCAM is based on the concept of gradient-based CAM. It leverages the gradients of the final convolutional layer with respect to the predicted class to understand which parts of the input image are crucial for making that prediction. Around 33\% of the studies (20 out of 61) utilized Grad-CAM for visualizing the final predictions. Another technique, namely Grad-CAM++ enhances Grad-CAM by substituting the globally averaged gradients with a weighted average of the gradients at the pixel level. This adaptation takes into account the significance of individual pixels in influencing the final prediction, resulting in more effective visual interpretations of CNN model predictions. Grad-CAM++ effectively overcomes the limitations of Grad-CAM, particularly in scenarios involving multiple instances of a class in an image.

\noindent
\textbf{Eigen class activation map} (Eigen-CAM) is a variation of CAM that incorporates the use of principal components of the learned convolutional activations \citep{Muhammad2020eigen}. It offers more accurate localization of important regions in an image and provides a deeper understanding of the underlying features. \citet{Bany2021} employed Eigen-CAM as a tool for localizing osteoarthritis (OA) features in X-ray images, using the Kellgren Lawrence grading scheme. The application of Eigen-CAM revealed significant findings, specifically highlighting the medial and lateral margins of the knee joint. These highlighted regions correspond to joint space narrowing and osteophytes sign, offering valuable insights into the presence and severity of OA-related changes in the knee joint.

\paragraph{\textbf{DeepLIFT}}
DeepLIFT was used by \citet{Chan2021} to quantitatively assess the contribution of each risk factor to the model's prediction. The assessment was carried out by computing the relative backpropagated gradients of the risk factors with regard to the model's prediction output. Their analysis revealed that for the prediction of knee OA onset, the medial JSN exhibited the highest DeepLIFT gradient, followed by history of injury. However, in the prediction of knee OA deterioration, diabetes and smoking habits showed the second and third highest gradients, respectively, alongside medial JSN, indicating their greater impact compared to injury.

\subsubsection{Game theory based}
SHapley Additive Explanations (SHAP) is a widely favoured post-hoc approach for handling tabular data in machine learning models. This approach is rooted in game theory that provides local explanations for individual predictions in the models. By calculating Shapley Values, it assigns importance values to each feature based on their interactions and contributions to the prediction outcome. It enables a comprehensive understanding of the factors driving each prediction and facilitates interpretability by identifying the most influential features in the decision-making process. The findings of all eight studies that utilized SHAP on tabular data were summarized in Table \ref{tab:game_theory_based}.

\subsubsection{Case based}
Case-Based approach is a knowledge-driven approach in which all relevant knowledge is pre-programmed and explicitly specified. \citet{Esteves2017case} demonstrated a case-based methodology for detecting knee osteoarthritis. Their proposed methodology integrated a logic programming approach to knowledge representation and reasoning with a case-based approach to computing, resulting in a comprehensive framework for effective problem-solving in OA field.

\subsubsection{Knowledge extraction based}
\begin{figure*}
\centering
\includegraphics[width=.75\linewidth]{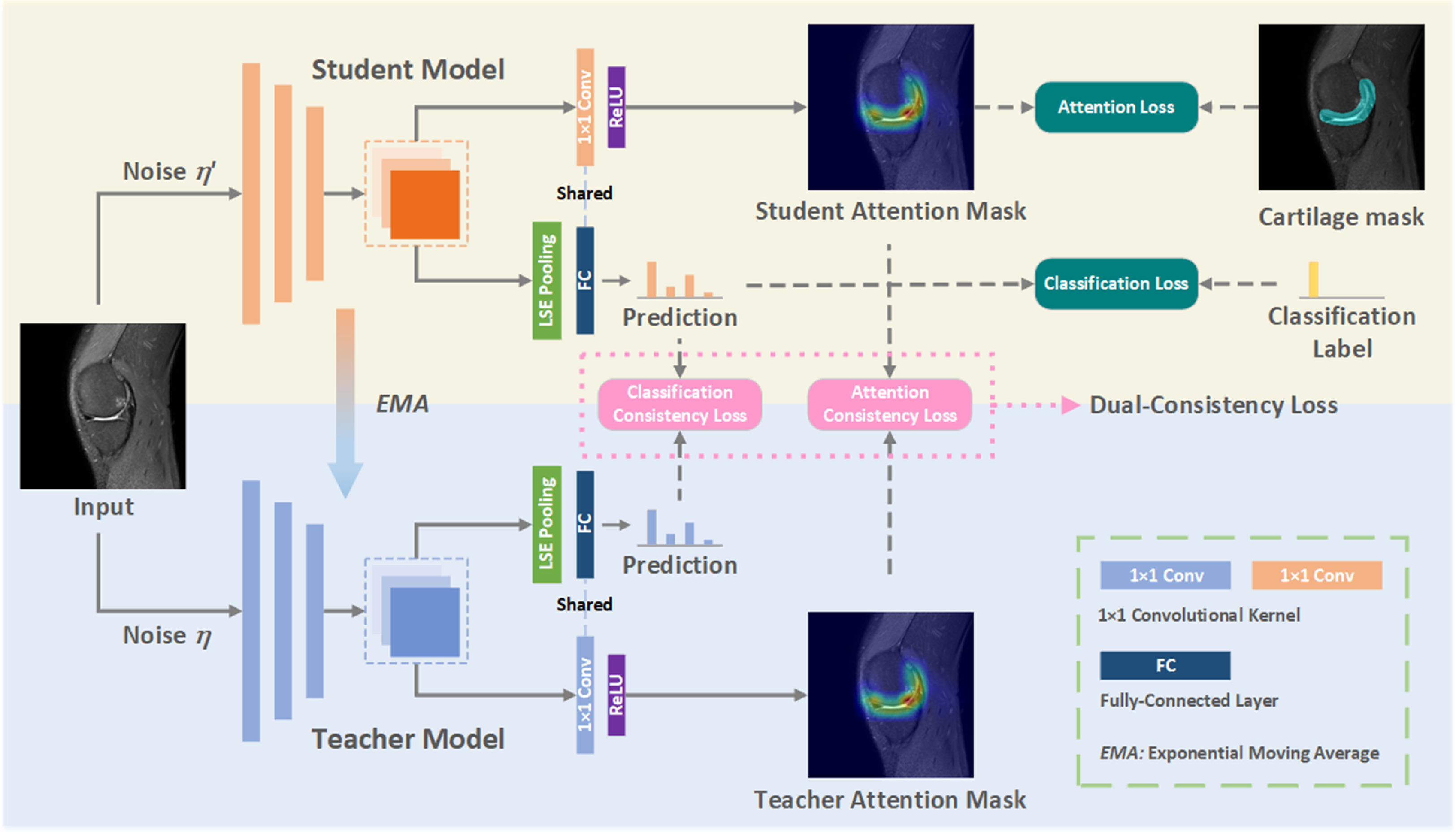}
\caption{Example of knowledge distillation. Adapted from \citep{Huo2022automatic}}\label{fig_knowledge_distillation}
\end{figure*}

Knowledge distillation is the core of the knowledge extraction based approaches. \citet{Huo2022automatic} demonstrated the use of dual-consistency mean teacher model (Figure \ref{fig_knowledge_distillation}) to discriminate cartilage damages. Both the teacher sub-model and the student sub-model shared common network architecture, but the teacher model utilized an exponential moving average (EMA) strategy for weight updates. This approach involved averaging the student network's weights across multiple training steps, enabling the teacher model to maintain consistent predictions and effectively guide the student network, particularly for unlabelled data. Recent study by \citet{Aladhadh2023knee} employed knowledge distillation to convey pixel and pair-wise information from a teacher network to a student network. The teacher network, built upon HRNet-W, featured a head convolution layer consisting of 64 filters and a 3×3 kernel, whereas the student network was equipped with 32 filters and 3×3 kernels. The student network was trained using pixel-wise knowledge extracted from heatmaps generated by the more complex teacher network with loss function as shown in Equation \ref{equation_loss_function_Aladhadh}, enabling the student network to adopt a simpler and more compact architecture. 
\begin{equation}
\label{equation_loss_function_Aladhadh}
    L_{pi}=\frac {\sum \nolimits _{i\in \Re } KL(h_{i}^{s}\vert \vert h_{i}^{t}) }{\hat {w}\times \hat {h}},\quad \Re =1,2,\ldots,\hat {\text {w}}\times \hat {\text {h}}
\end{equation}
\noindent
where ${\displaystyle \ h_{i}^{s}}$ represents the response of the pixel at ${\displaystyle \ ith}$ position in student network, ${\displaystyle \ h_{i}^{t}}$ represents the response by teacher network at ${\displaystyle \ ith}$ position of pixel, ${\displaystyle \ KL}$ represents the Kullback–Leibler exhibiting divergence among two heatmaps, and ${\displaystyle \ \hat {w}\times \hat {h}}$ represents feature map.

\citet{Kornreich2022combining} presented a novel two-stage method inspired by multiple instance learning. This method aimed to identify regions of high likelihood for pathologies by leveraging mixed-format data, which encompassed categorical and positional labels. Their approach incorporated a UNet network along with a morphological peak-finding algorithm to accurately localize defects. Prior to pathology detection, the images were automatically cropped around the anterior cruciate ligament or medial compartment cartilage. Additionally, they employed a deep reinforcement learning model to detect two anatomical landmarks, namely the intercondylar eminence and the fibular styloid, which were used to position a volume of interest in relation to the location of these landmarks.

\subsubsection{Neural based}
\begin{figure*}
\centering
\includegraphics[width=0.65\linewidth]{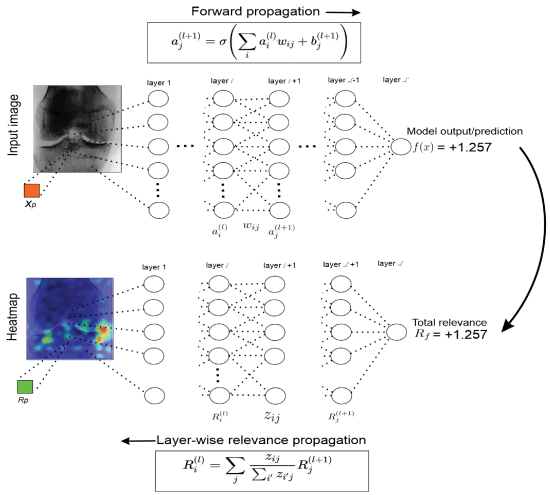}
\caption{Example of layer-wise relevance propagation (LRP) for knee OA detection. Adapted from \citep{Karim2021}}\label{fig_layer_wise_relevance_propagation}
\end{figure*}

Neural-based techniques encompass methods that explain specific predictions, simplify neural networks, and visualize the features and concepts learned by the network. \citet{Ciliberti2022} conducted feature important analysis on a pre-developed model based on random forest algorithm. Their findings demonstrated that cartilage and bone features, including the volume of femoral cartilage and patellar density, played a significant role in classifying the status of the knee, whether it was healthy, degenerative, or traumatic. \citet{Karim2021} implemented another neural based technique, namely layer-wise relevance propagation (LRP) as illustrated in Figure \ref{fig_layer_wise_relevance_propagation} to tackle important pixels by running a forward pass through the neural network. In addition, deep Taylor decomposition (DTD) was utilized to backpropagate the relevance ${\displaystyle R_{t}^{(L)}}$, allowing for the generation of a visualizable relevance map ${\displaystyle R_{LRP}}$.

\begin{table*}[]
\caption{Summary of attribution-based backpropagation XAI techniques from included papers. AUC: area under curve; BMI: body mass index; CAM: class activation mapping; CBAM: convolutional block attention module; GAP: global average pooling; GradCAM: gradient-weighted class activation mapping; Grad-CAM++: gradient-guided class activation maps; JSN: joint space narrowing; KOOS: knee injury and osteoarthritis outcome; LRP: layer-wise relevance propagation; MSE: mean squared error; ROC: receiver operating characteristic; TKR: total knee replacement.}\label{tab_attribution_based_XAI}
\resizebox{\textwidth}{!}{%
\begin{tabularx}{\textwidth} {p{1.0cm} p{1.25cm} p{1.5cm} p{2.0cm}  p{1.5cm} p{3.0cm} p{5.25cm}}
\toprule
\textbf{Paper} & \textbf{XAI method} & \textbf{Type of data} & \textbf{Evaluated model (performance)} & \textbf{Target} & \textbf{XAI findings} & \textbf{Visualization of XAI}\\
\midrule

\citet{Tiulpin2018}     
& GradCAM
& Imaging - X-ray
& Baseline: ResNet-34 (67.49\% accuracy, 0.83 Kappa coefficient, 0.51 MSE)
& Classification of OA severity based on KL grades \par (5 classes)
& Joint centre area was highlighted. 
& Baseline: 
\par
\raisebox{0.1\totalheight}{\includegraphics[width=0.21\textwidth]{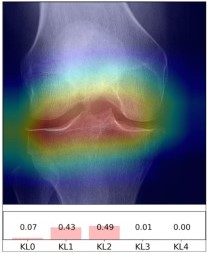}}
\\
&&
& Proposed: Ensemble method with SE-Resnet50 and SE-ResNext50-32x4d (66.71\% accuracy, 0.83 Kappa coefficient, 0.48 MSE)
&&
Areas exhibiting osteophytes were highlighted.
&
Proposed model: 
\par
\raisebox{0.1\totalheight}{\includegraphics[width=0.21\textwidth]{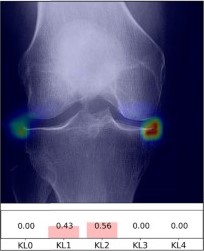}}\\
\hline

\citet{Tiulpin2019}     
& GradCAM
& Multimodal: imaging - X-ray and clinical data
& Deep CNN with Gradient Boosting Machine (0.79 AUC, 0.68 average precision)
& Prediction of change in KL grade in 60 months \par (3 classes)
& The GradCAM attention maps consistently emphasized the presence of tibial spines.
& \raisebox{-\totalheight}{\includegraphics[width=0.21\textwidth]{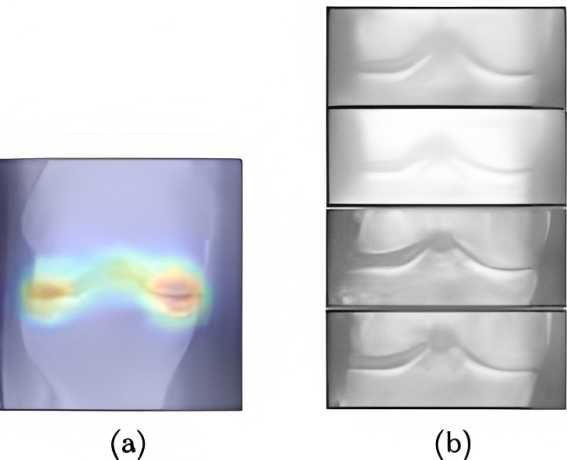}} \raisebox{-\totalheight}{\includegraphics[width=0.21\textwidth]{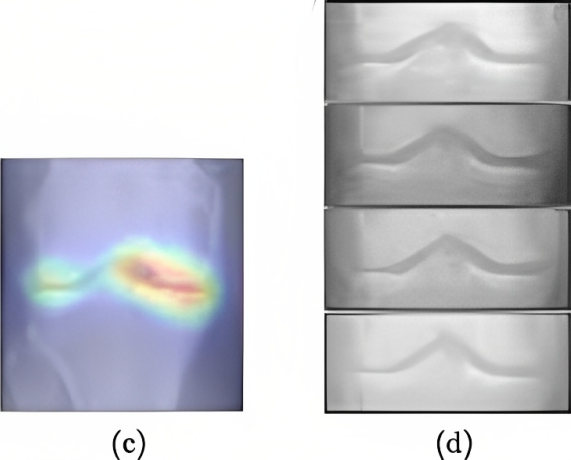}}
\\

\bottomrule
\end{tabularx}}
\end{table*}

\begin{table*}[]
\ContinuedFloat
\caption{(\textit{continued}).}
\resizebox{\textwidth}{!}{%
\begin{tabularx}{\textwidth} {p{1.0cm} p{1.25cm} p{1.5cm} p{2.0cm}  p{1.5cm} p{3.0cm} p{5.25cm}}
\toprule
\textbf{Paper} & \textbf{XAI method} & \textbf{Type of data} & \textbf{Evaluated model (performance)} & \textbf{Target} & \textbf{XAI findings} & \textbf{Visualization of XAI}\\
\midrule

\citet{Chen2019fully}
& GradCAM
& Imaging - X-ray
& VGG19 with adjustable ordinal loss (69.7\% classification accuracy and 0.344 mean absolute error)
& Classification of OA severity based on KL grades \par (5 classes)
& The localization of key indicators of knee OA, including JSN, subchondral sclerosis, and osteophyte formation were achieved.
& \raisebox{-\totalheight}{\includegraphics[width=0.20\textwidth]{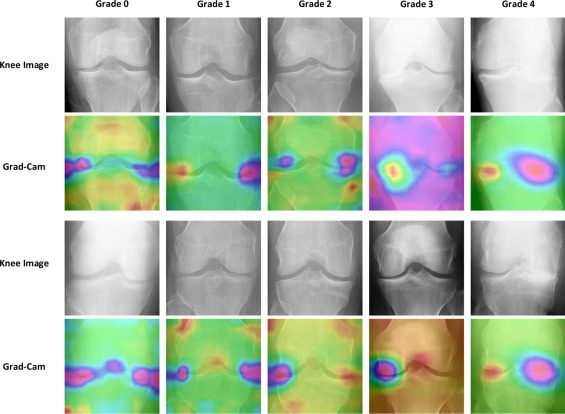}}\\
\hline

\citet{Norman2019applying}
& Saliency maps
& Imaging - X-ray
& DenseNet (Sensitivity rates of no OA, mild, moderate, and severe OA were 83.7, 70.2, 68.9, and 86.0\% respectively. Specificity rates were 86.1, 83.8, 97.1, and 99.1\%.)
& Classification of OA severity based on KL grades \par (4 classes)
& Saliency map accurately captured osteophytes in mild OA and JSN in severe OA. However, misclassifications could occur, such as assigning high importance to a screw instead of joint features in a moderate OA case.
& \raisebox{-\totalheight}{\includegraphics[width=0.20\textwidth]{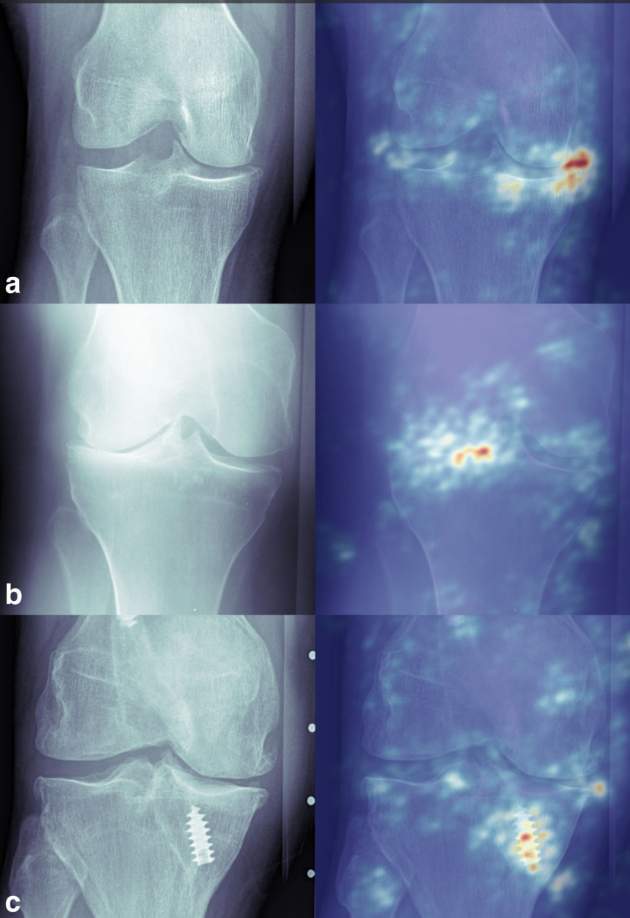}}\\
\hline

\citet{Leung2020}     
& GradCAM
& Imaging - X-ray
& Multitask ResNet34 model trained with transfer learning (0.87 AUC)
& Prediction of TKR within 9 years \par (2 classes) \par Classification of OA severity based on KL grades \par (5 classes)
& The findings demonstrated that the model made predictions based on regions located near the knee joint space for classification of samples between the KL0 and KL2 grades.
& \raisebox{-\totalheight}{\includegraphics[width=0.22\textwidth]{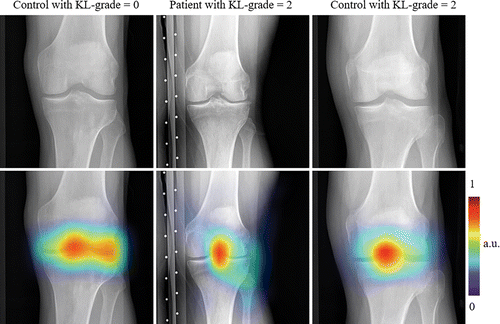}}\\
\hline

\citet{Kim2020}     
& GradCAM
& Imaging - X-ray and clinical information
& SE-ResNet (AUC of 0.97 for KL0, 0.85 for KL1, 0.75 for KL2, 0.86 for KL3, and 0.95 for KL4)
& Classification of OA severity based on KL grades \par (5 classes)
& The model managed to detect knee joint in raw image data of KL 0, 3, and 4 without cropping the images, but it struggled to detect knee joint of raw images with KL 1 and 2.
& \raisebox{-\totalheight}{\includegraphics[width=0.21\textwidth]{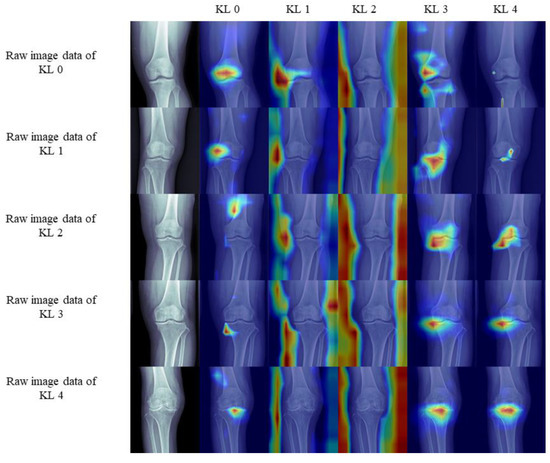}}
\\

\bottomrule
\end{tabularx}}
\end{table*}

\begin{table*}[]
\ContinuedFloat
\caption{(\textit{continued}).}
\resizebox{\textwidth}{!}{%
\begin{tabularx}{\textwidth} {p{1.0cm} p{1.25cm} p{1.5cm} p{2.0cm}  p{1.5cm} p{3.0cm} p{5.25cm}}
\toprule
\textbf{Paper} & \textbf{XAI method} & \textbf{Type of data} & \textbf{Evaluated model (performance)} & \textbf{Target} & \textbf{XAI findings} & \textbf{Visualization of XAI}\\
\midrule

\citet{Chang2020}     
& CAM
& Imaging - MRI
& Siamese neural network with six convolutional layers in each network (75.70\% accuracy)
& Prediction of unilateral knee pain based on WOMAC \par (2 classes)
& Effusion or synovitis (c, d) was identified as the most prevalent structural abnormality associated with frequent knee pain in 95 out of 107 subjects (88.8\%), followed by bone marrow lesion (5.6\%), Hoffa fat pad abnormalities (3.7\%), cartilage loss (1.9\%), and meniscal damage (0.93\%).
& \raisebox{-\totalheight}{\includegraphics[width=0.22\textwidth]{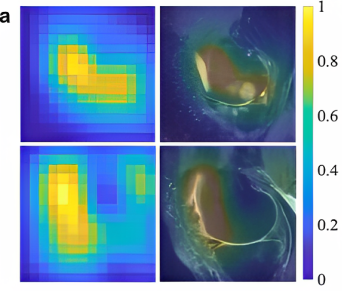}} \raisebox{-\totalheight}{\includegraphics[width=0.22\textwidth]{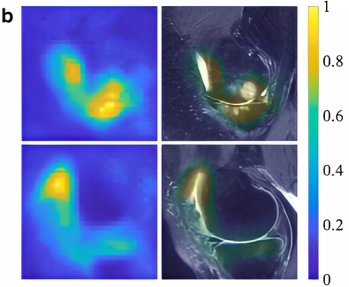}} \raisebox{-\totalheight}{\includegraphics[width=0.20\textwidth]{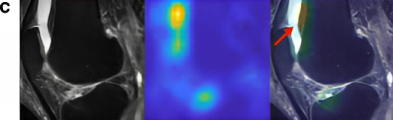}} \raisebox{-\totalheight}{\includegraphics[width=0.20\textwidth]{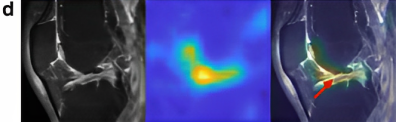}}
\\
\hline

\citet{Zhang2020}     
& GradCAM
& Imaging - X-ray
& ResNet34 with CBAM (74.81\% multi-class average accuracy, 0.36 mean squared error, and 0.88 quadratic Kappa score)
& Classification of OA severity based on KL grades \par (5 classes)
& With CBAM, the model focused on the center, medial, and lateral parts of the knee for predicting KL1 and KL0. ResNet34 with CBAM exhibited a more focused attention on the specific regions that were already highlighted by ResNet34 without CBAM.
& \raisebox{-\totalheight}{\includegraphics[width=0.21\textwidth]{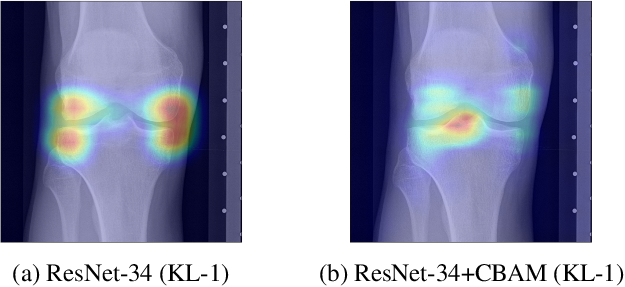}} \raisebox{-\totalheight}{\includegraphics[width=0.21\textwidth]{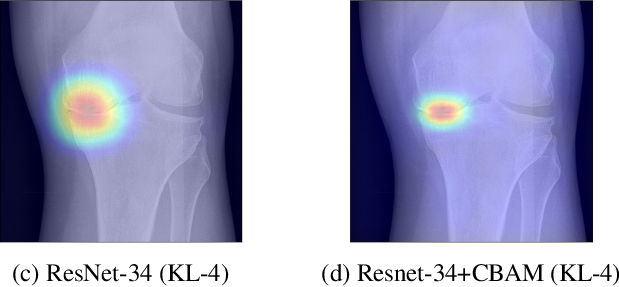}}\\
\\

\bottomrule
\end{tabularx}}
\end{table*}

\begin{table*}[]
\ContinuedFloat
\caption{(\textit{continued}).}
\resizebox{\textwidth}{!}{%
\begin{tabularx}{\textwidth} {p{1.0cm} p{1.25cm} p{1.5cm} p{2.0cm}  p{1.5cm} p{3.0cm} p{5.25cm}}
\toprule
\textbf{Paper} & \textbf{XAI method} & \textbf{Type of data} & \textbf{Evaluated model (performance)} & \textbf{Target} & \textbf{XAI findings} & \textbf{Visualization of XAI}\\
\midrule

\citet{Thomas2020}     
& Saliency maps
& Imaging - X-ray
& DenseNet169 (0.70 F1, 0.71 accuracy, 0,86 Cohen weighted kappa)
& Classification of OA severity based on KL grades \par (5 classes)
& Osteophyte formation sites demonstrated high influence to final KL prediction.
& \raisebox{-\totalheight}{\includegraphics[width=0.21\textwidth]{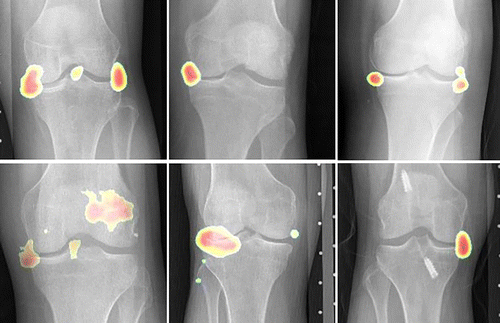}}\\
\hline

\citet{Bany2021}     
& Eigen-CAM
& Imaging - X-ray
& Stacked ensemble learning using CNN with SVM as super learner
& Classification of OA severity based on KL grade \par (5 classes)
& OA features (JSN and Osteophytes) in the joint medial and lateral margins
& \raisebox{-\totalheight}{\includegraphics[width=0.21\textwidth]{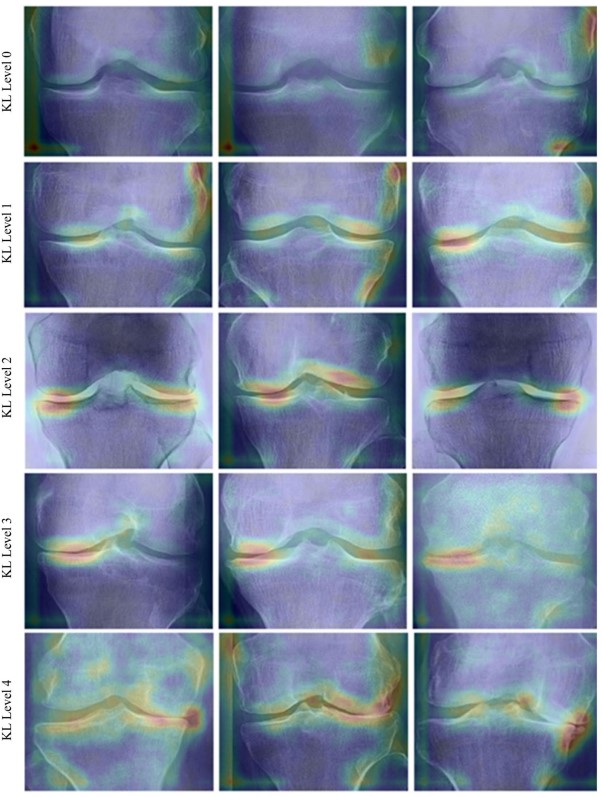}}\\
\hline

\citet{Morales2021}     
& Spherical GradCAM
& Imaging - MRI
& ResNet50 (53.7 - 58.8 sensitivity, 67.4 - 82.1 specificity, 68.1 - 74.4 AUC) 
& Detection of meniscal tears \par (2 classes)
& The similar patterns observed in both the true positive and true negative groups for the femur bone shape feature suggested that the model exploited similar features for assessing pain presence and absence.
& \raisebox{-\totalheight}{\includegraphics[width=0.21\textwidth]{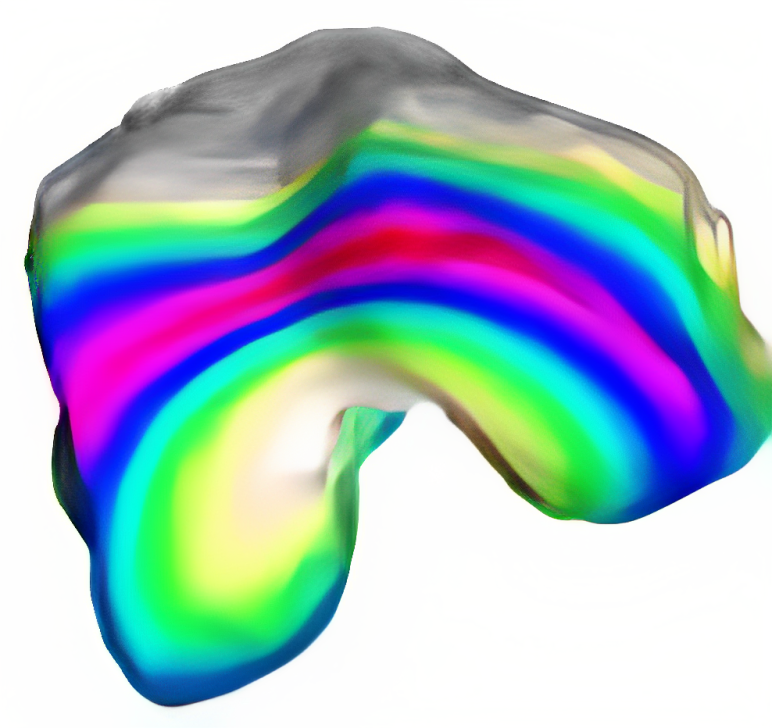}}
\\

\bottomrule
\end{tabularx}}
\end{table*}

\begin{table*}[]
\ContinuedFloat
\caption{(\textit{continued}).}
\resizebox{\textwidth}{!}{%
\begin{tabularx}{\textwidth} {p{1.0cm} p{1.25cm} p{1.5cm} p{2.0cm}  p{1.5cm} p{3.0cm} p{5.25cm}}
\toprule
\textbf{Paper} & \textbf{XAI method} & \textbf{Type of data} & \textbf{Evaluated model (performance)} & \textbf{Target} & \textbf{XAI findings} & \textbf{Visualization of XAI}\\
\midrule

\citet{Moon2021}     
& GradCAM
& Imaging - weight bearing whole-leg X-ray
& CNN architecture consisted of six stacked SE-ResNet blocks, followed by Log-Sum-Exp pooling, a fully connected layer, and Softmax activation functions (95.1\% cumulative score, 0.054 mean absolute error)
& Classification of WBL ratio \par (7 classes)
& For WBL ratios of 0.2 to 0.6, heatmap signals were concentrated around the knee joint area. However, for a WBL ratio of 0.0, the heatmap signal appeared at multiple points, including the femoral diaphysis, metaphysis, fibular head, and tibial diaphysis. For a WBL ratio of 0.1, the heat map signal specifically appeared on the tibial diaphysis.
& \raisebox{-\totalheight}{\includegraphics[width=0.21\textwidth]{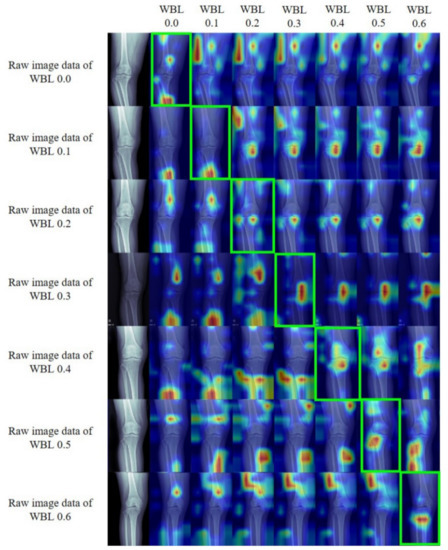}}\\
\hline

\citet{Schiratti2021}     
& GradCAM
& Imaging - MRI and clinical variables
& EfficientNet-B0 network with attention sub-model and classification sub-model (task 1 - 65\% ROC-AUC, 13\% precision, 84\% recall; task 2 - 66.8\% mean precision-recall AUC, 72.4\% mean ROC-AUC, 65.2\% mean weighted F1)
& Prediction of JSN progression at 12 months \par (2 classes) \par Prediction of pain severity \par (2 classes)
& For JSN progression, the medial joint space was found to be highly relevant, while for pain prediction, the intra-articular space emerged as a significant factor.
& \raisebox{-\totalheight}{\includegraphics[width=0.21\textwidth]{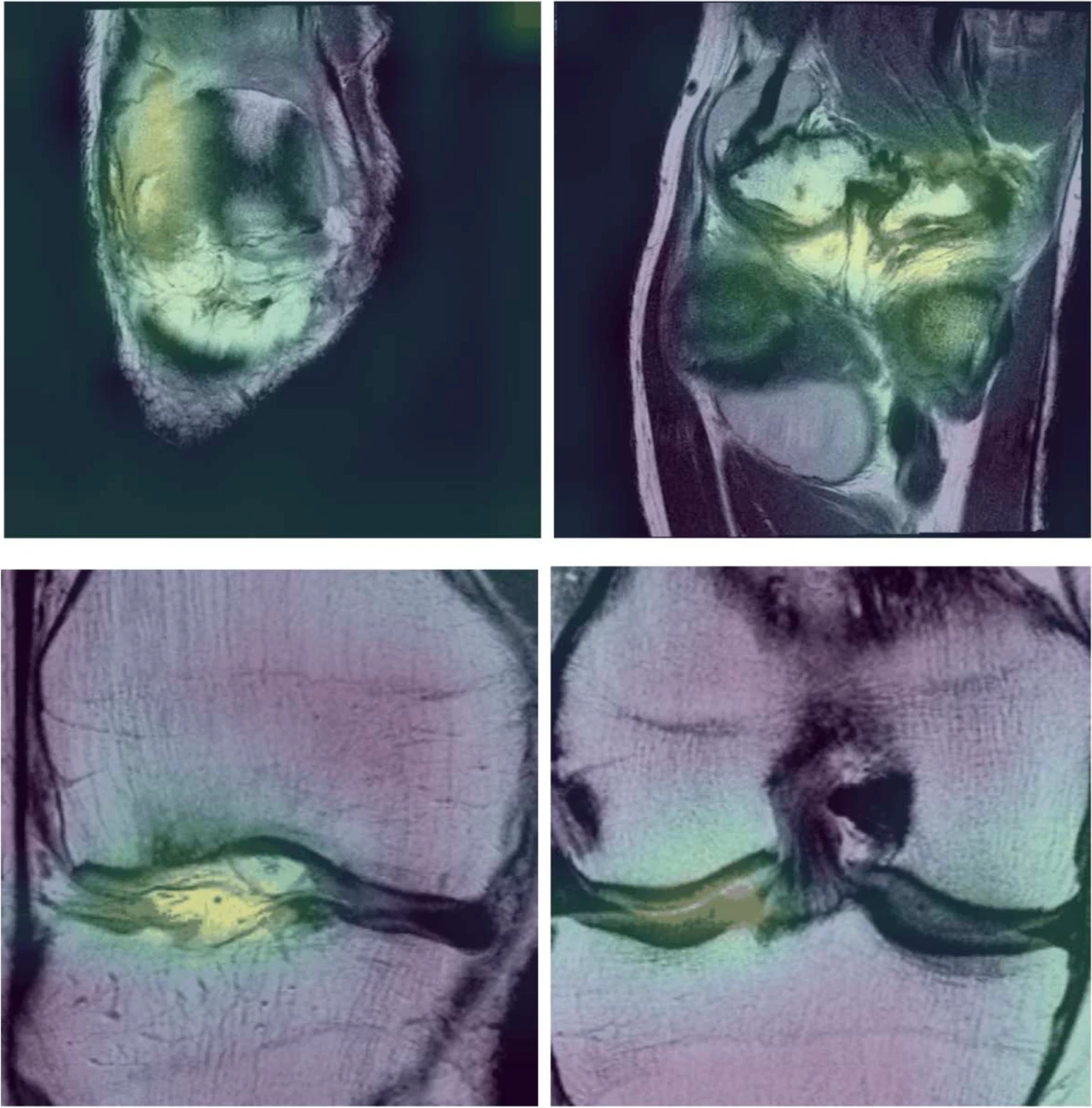}}\\
\hline

\citet{Zeng2021}    
& GradCAM
& Imaging - X-ray
& CNN
& Presence of OA \par (2 classes) \par Classification of OA severity based on KL grade \par (5 classes)
& No further explanation by the authors.
& \raisebox{-\totalheight}{\includegraphics[width=0.21\textwidth]{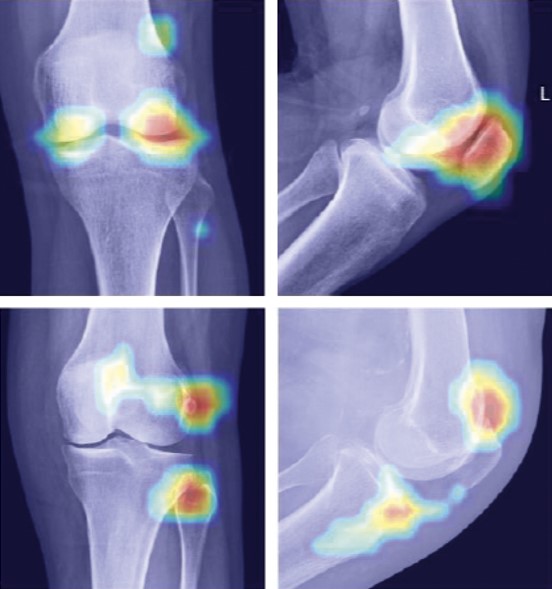}}
\\

\bottomrule
\end{tabularx}}
\end{table*}

\begin{table*}[]
\ContinuedFloat
\caption{(\textit{continued}).}
\resizebox{\textwidth}{!}{%
\begin{tabularx}{\textwidth} {p{1.0cm} p{1.25cm} p{1.5cm} p{2.0cm}  p{1.5cm} p{3.0cm} p{5.25cm}}
\toprule
\textbf{Paper} & \textbf{XAI method} & \textbf{Type of data} & \textbf{Evaluated model (performance)} & \textbf{Target} & \textbf{XAI findings} & \textbf{Visualization of XAI}\\
\midrule

\citet{Karim2021}     
& GradCAM++ and LRP
& Imaging - X-ray, MRI
& DenseNet and VGG (91\% accuracy)
& OARSI JSN grades \par (4 classes)
& No further explanation by the authors
& \raisebox{-\totalheight}{\includegraphics[width=0.21\textwidth]{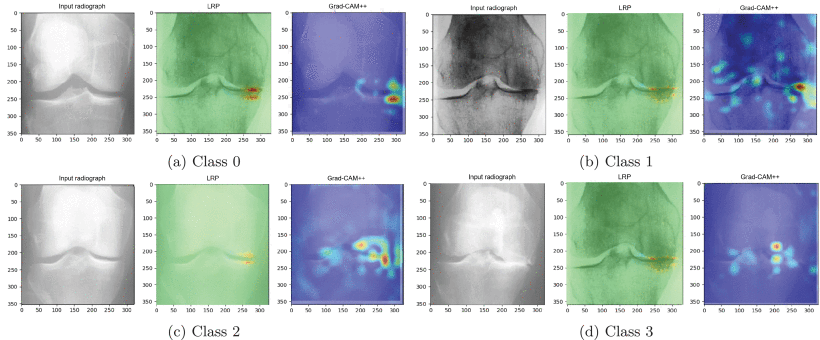}}
\raisebox{-\totalheight}{\includegraphics[width=0.21\textwidth]{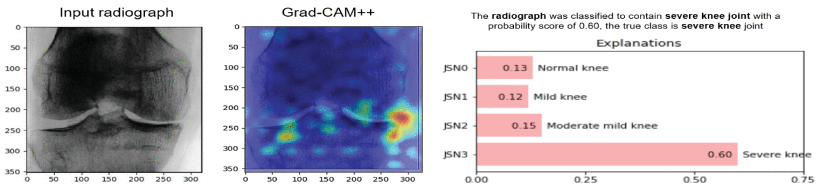}}
\raisebox{-\totalheight}{\includegraphics[width=0.21\textwidth]{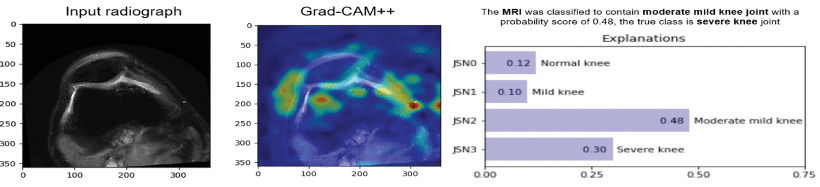}}
\raisebox{-\totalheight}{\includegraphics[width=0.21\textwidth]{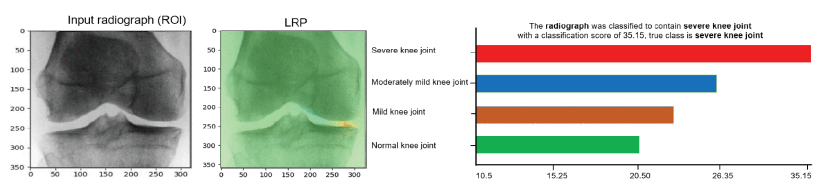}}
\raisebox{0.1\totalheight}{\includegraphics[width=0.21\textwidth]{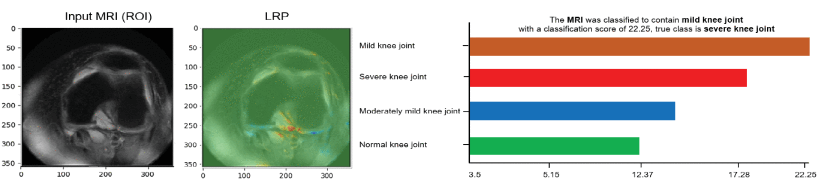}}\\
\hline

\citet{Wang2021automatic}     
& GradCAM
& Imaging - X-ray
& YOLO, ResNet50 backbone with visual transformer (69.18\% accuracy)
& Classification of OA severity based on KL grades \par (5 classes)
& Unlike ResNet50, which showed a single centralized high attention region, the proposed method exhibited high-weighted areas spread across both sides of the X-ray images, leveraging the correlation between small regions. Additionally, the proposed method outperformed ResNet50 in locating JSN and detecting lesions on the medial or lateral edge of the femur, including sclerosis or bone spurs.
& \raisebox{-\totalheight}{\includegraphics[width=0.21\textwidth]{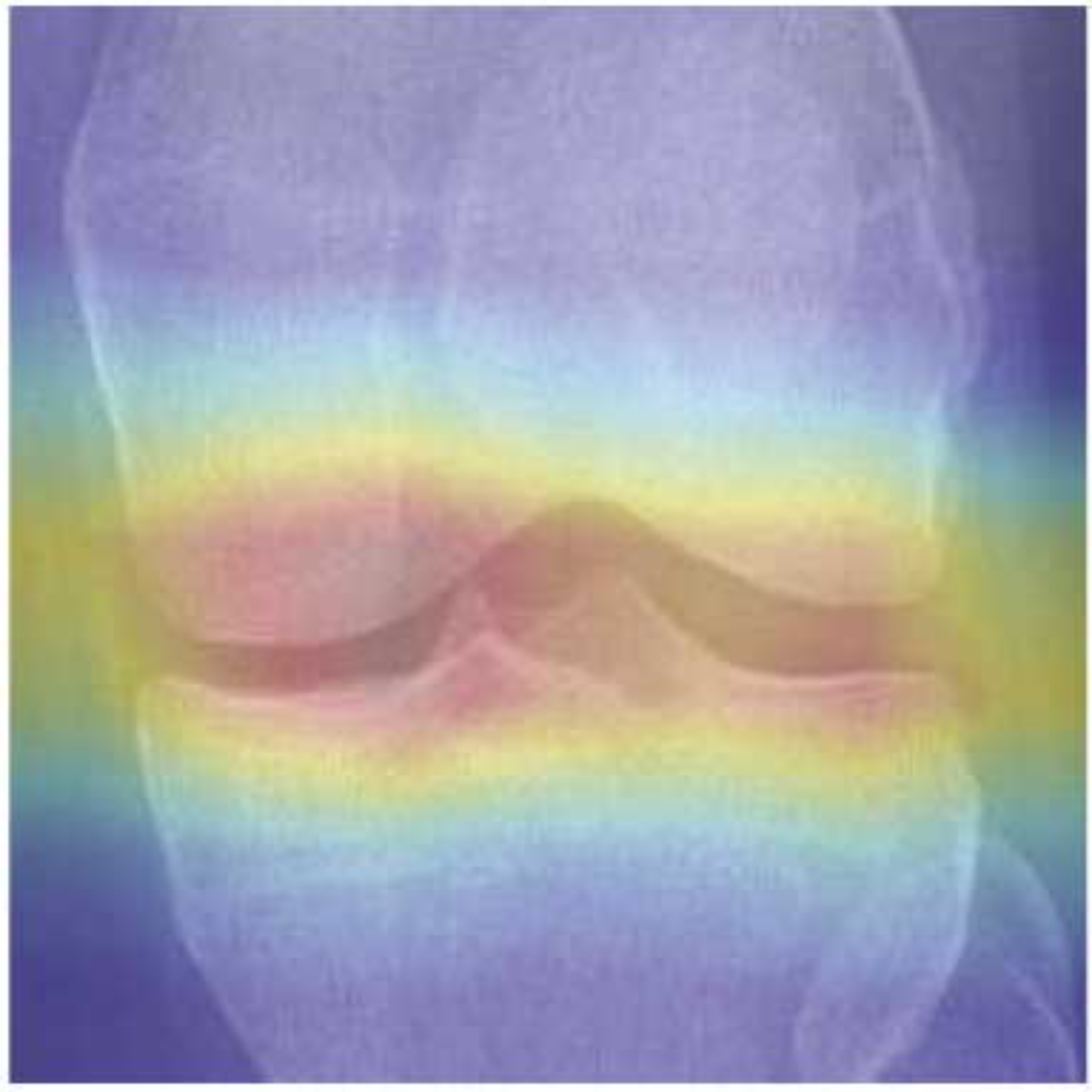}} 
\par (a) ResNet50 
\par \raisebox{-\totalheight}{\includegraphics[width=0.21\textwidth]{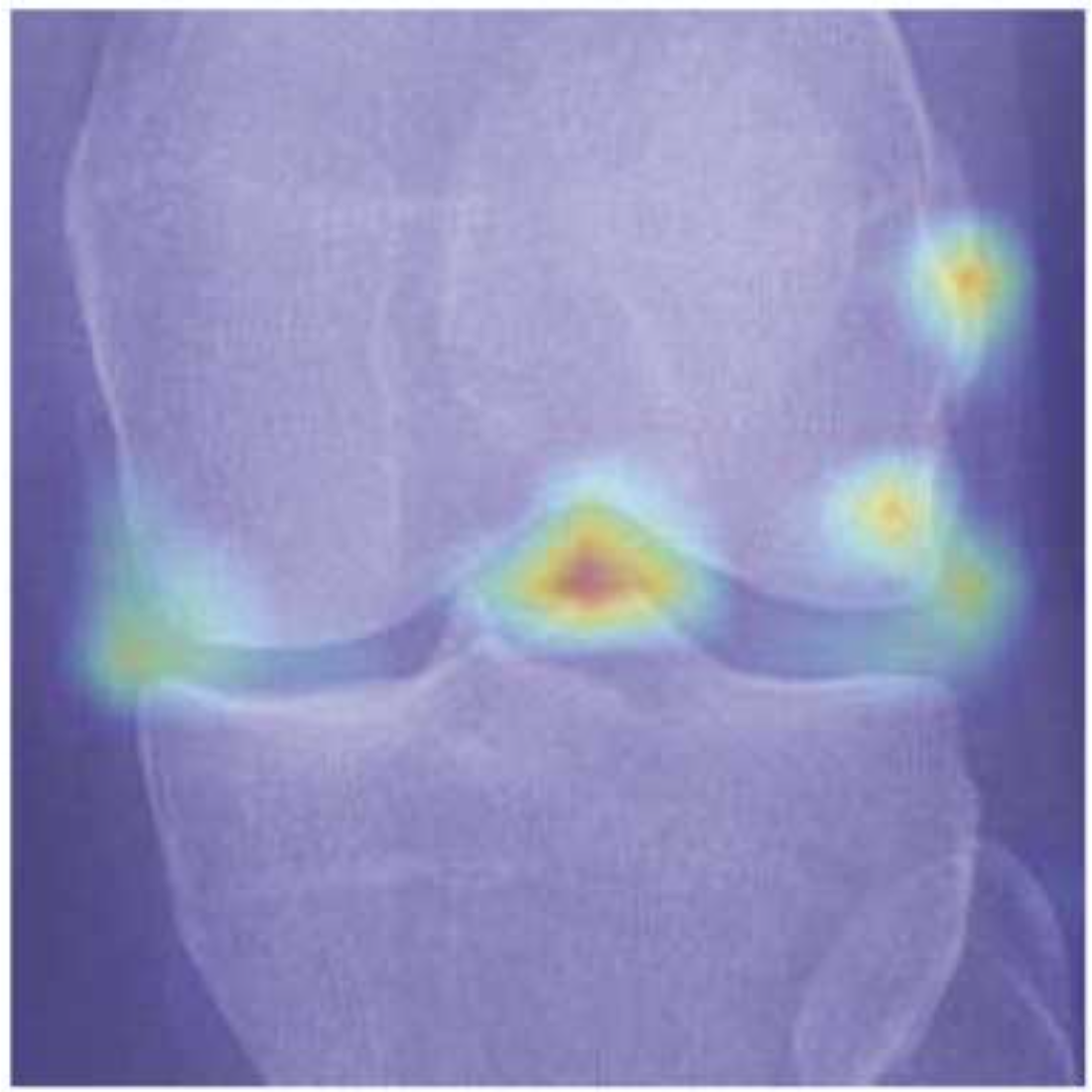}} 
\par (b) With visual transformer
\\
\hline

\citet{Olsson2021}
& Integrated gradient heatmap
& Imaging - X-ray
& ResNet (overall AUC > 0.80 with highest AUC for KL0 with an AUC of 0.97, with sensitivity and specificity of 97 and 88\%)
& Classification of OA severity based on KL grades \par (5 classes)
& For wrong predictions, heatmap activity was focused on the implant, suggesting that the network was responding to persistent indications of a previously treated medial arthrosis.
& \raisebox{-\totalheight}{\includegraphics[width=0.21\textwidth]{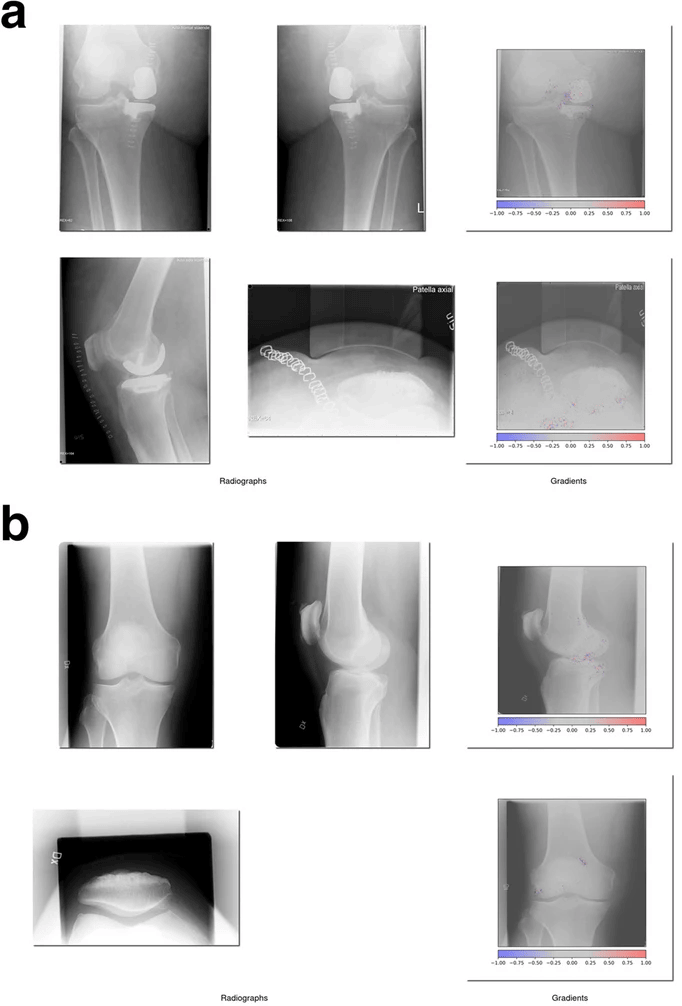}} 
\\

\bottomrule
\end{tabularx}}
\end{table*}

\begin{table*}[]
\ContinuedFloat
\caption{(\textit{continued}).}
\resizebox{\textwidth}{!}{%
\begin{tabularx}{\textwidth} {p{1.0cm} p{1.25cm} p{1.5cm} p{2.0cm}  p{1.5cm} p{3.0cm} p{5.25cm}}
\toprule
\textbf{Paper} & \textbf{XAI method} & \textbf{Type of data} & \textbf{Evaluated model (performance)} & \textbf{Target} & \textbf{XAI findings} & \textbf{Visualization of XAI}\\
\midrule

\citet{Feng2021automated}     
& GradCAM
& Imaging - X-ray
& ResNet with CBAM and Mish activation function (70.23\% overall accuracy, 68.23\% recall, 70.25\% precision, and 67.55\% F1)
& Classification of OA severity based on KL grades \par (5 classes)
& No further explanation by the authors.
& \raisebox{-\totalheight}{\includegraphics[width=0.21\textwidth]{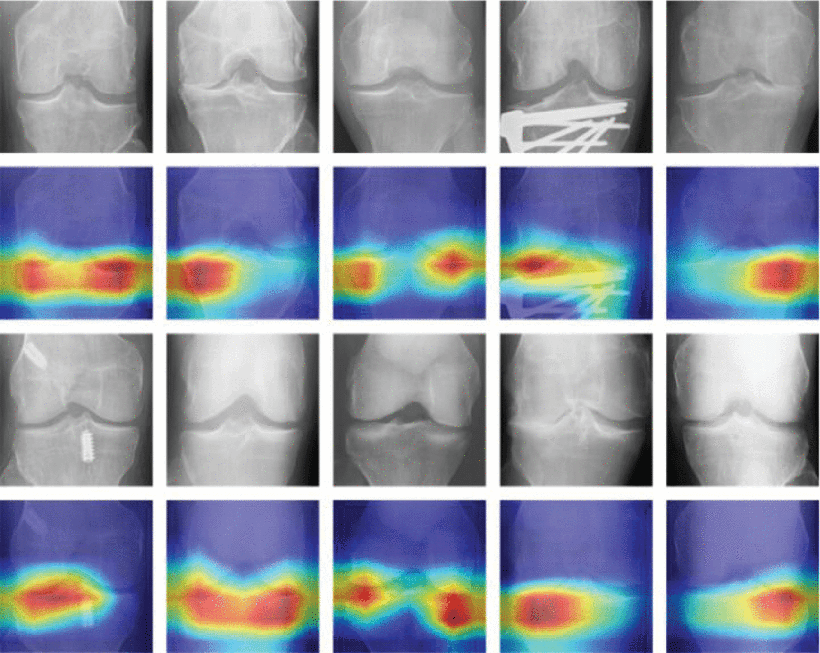}}
\\
\hline

\citet{Wang2021learning}    
& GradCAM
& Imaging - X-ray
& DenseNet121 with hybrid loss function using label confidence information (70.13\% and 87.42\% accuracies)
& Classification of OA severity based on KL grades \par (5 classes) 
\par Presence of early OA \par (2 classes)
& The model was incentivized to extract features from both the lateral and medial sides of the knee joint and provided more accurate predictions. However, there was issue of overestimating the severity in some cases.
& \raisebox{-\totalheight}{\includegraphics[width=0.21\textwidth]{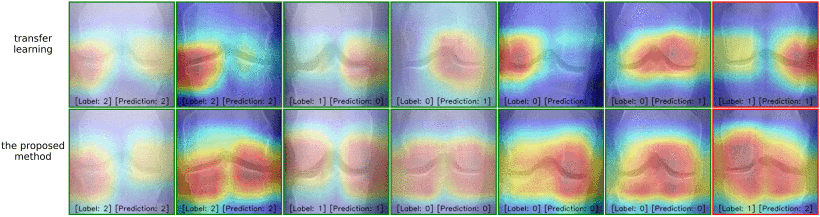}}
\\
\hline

\citet{Helwan2022update} 
& GradCAM
& Imaging - X-ray
& Wide residual network with transfer learning (72\% accuracy and 74\% precision)
& Classification of OA severity based on KL grades \par (5 classes)
& No further explanation by the authors.
& \raisebox{-\totalheight}{\includegraphics[width=0.21\textwidth]{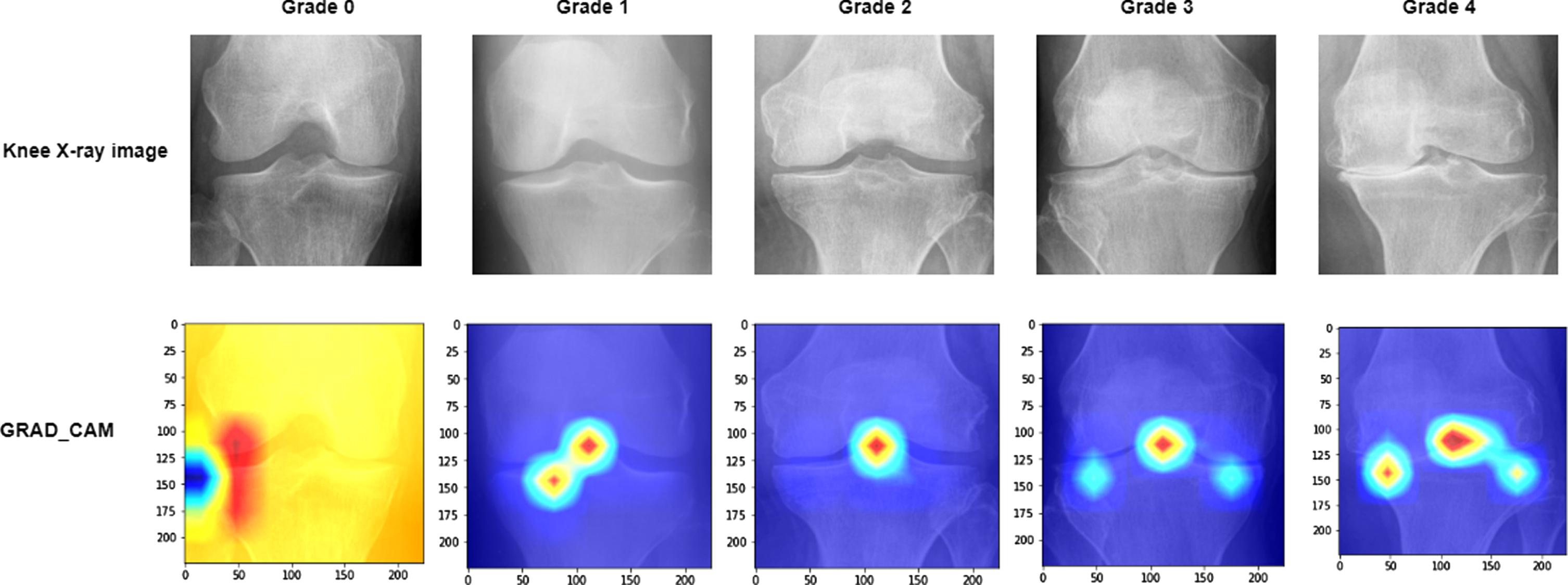}}
\\
\hline

\citet{Patron2022automatic}
& GradCAM
& Imaging - X-ray
& ResNeXt-50-32x4d with transfer learning (0.869 accuracy)
& Presence of tibial spiking \par (2 classes)
\par Presence of tibial spiking, including unsure ratings \par 3 classes
& In correct predictions, the heatmap highlighted the lateral tubercle in one case and showed no significant region in another non-spiking sample. In incorrect predictions, the model associated tibial spiking with the narrow appearance of the medial joint space in one case, and the presence of the medial tubercle influenced the spiking assessment in another case.
& \raisebox{-\totalheight}{\includegraphics[width=0.21\textwidth]{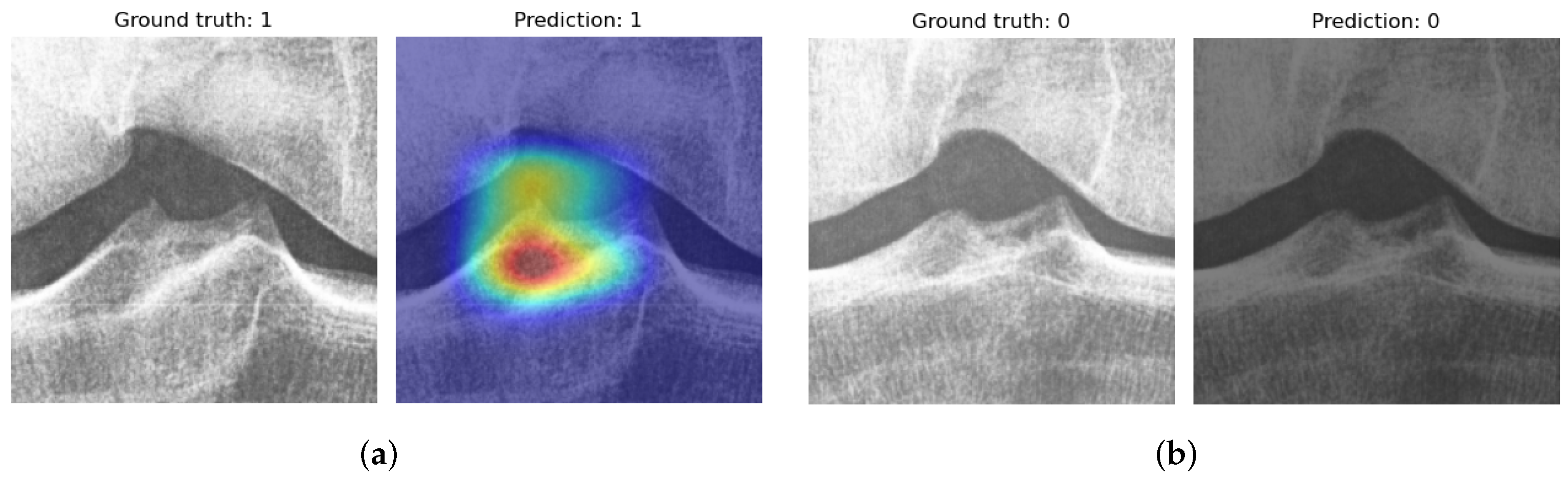}} \par (a) Correct prediction
\par \raisebox{-\totalheight}{\includegraphics[width=0.21\textwidth]{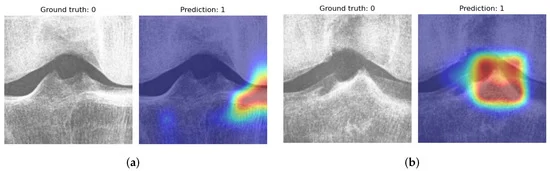}} \par (b) Wrong prediction
\\

\bottomrule
\end{tabularx}}
\end{table*}

\begin{table*}[]
\ContinuedFloat
\caption{(\textit{continued}).}
\resizebox{\textwidth}{!}{%
\begin{tabularx}{\textwidth} {p{1.0cm} p{1.25cm} p{1.5cm} p{2.0cm}  p{1.4cm} p{3.1cm} p{5.25cm}}
\toprule
\textbf{Paper} & \textbf{XAI method} & \textbf{Type of data} & \textbf{Evaluated model (performance)} & \textbf{Target} & \textbf{XAI findings} & \textbf{Visualization of XAI}\\
\midrule

\citet{Alshareef2022knee} 
& GradCAM
& Imaging - X-ray
& Vision Transformer (ViT) (71.2\% F1 and 70\% accuracy)
& Classification of OA severity based on KL grades \par (5 classes)
& No further explanation by the authors.
& \raisebox{-\totalheight}{\includegraphics[width=0.21\textwidth]{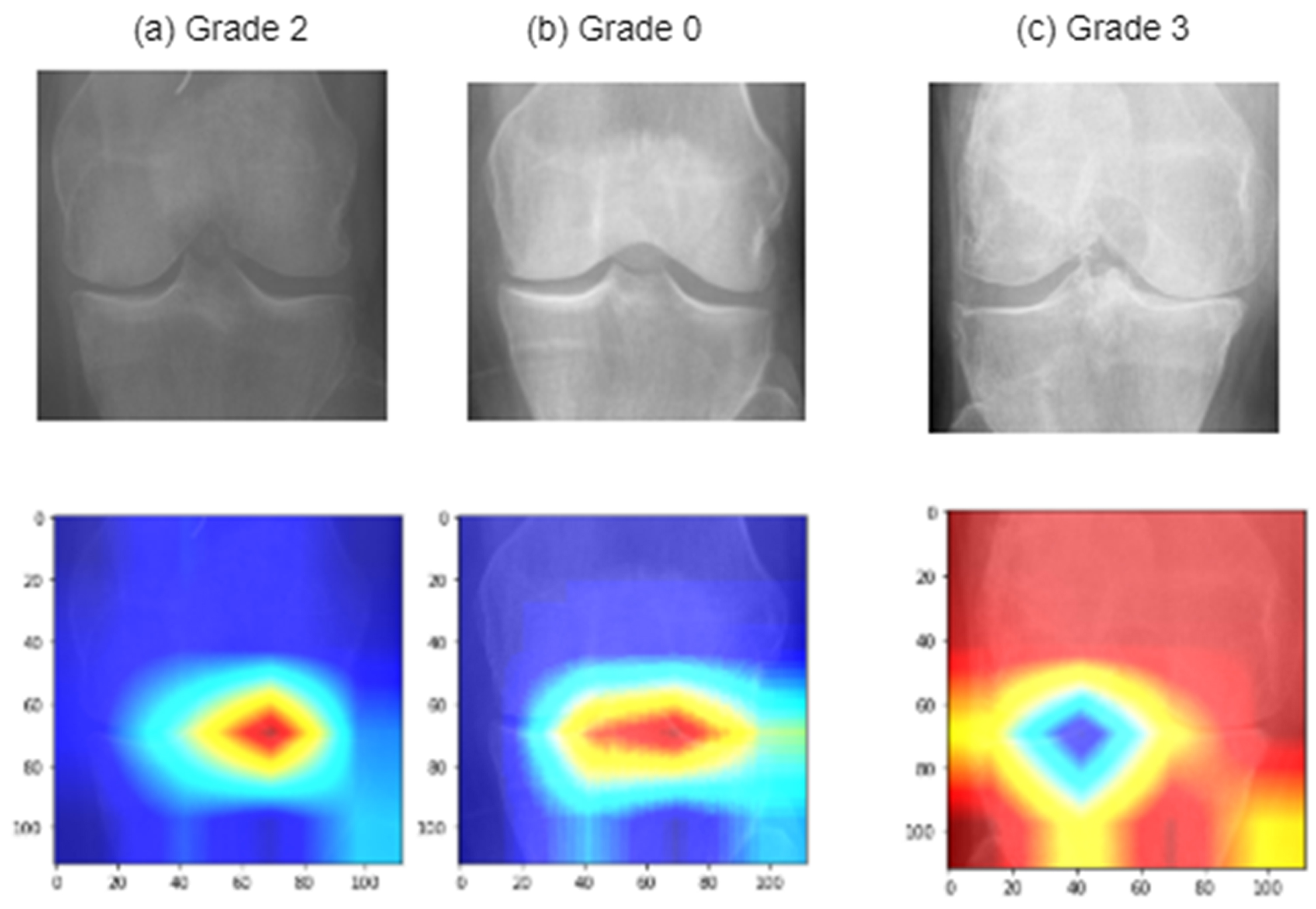}}
\\
\hline

\citet{Dunnhofer2022deep}    
& CAM
& Imaging - MRI
& MRNet (AlexNet) with MRPyrNet (composed of a Feature Pyramid Network with Pyramidal Detail Pooling) (0.834-0.974 ROC-AUC)
& Presence of ACL tear \par (2 classes)
\par Presence of meniscus tear \par (2 classes)
& The model was incentivized to extract features around joint centre.
& \raisebox{-\totalheight}{\includegraphics[width=0.21\textwidth]{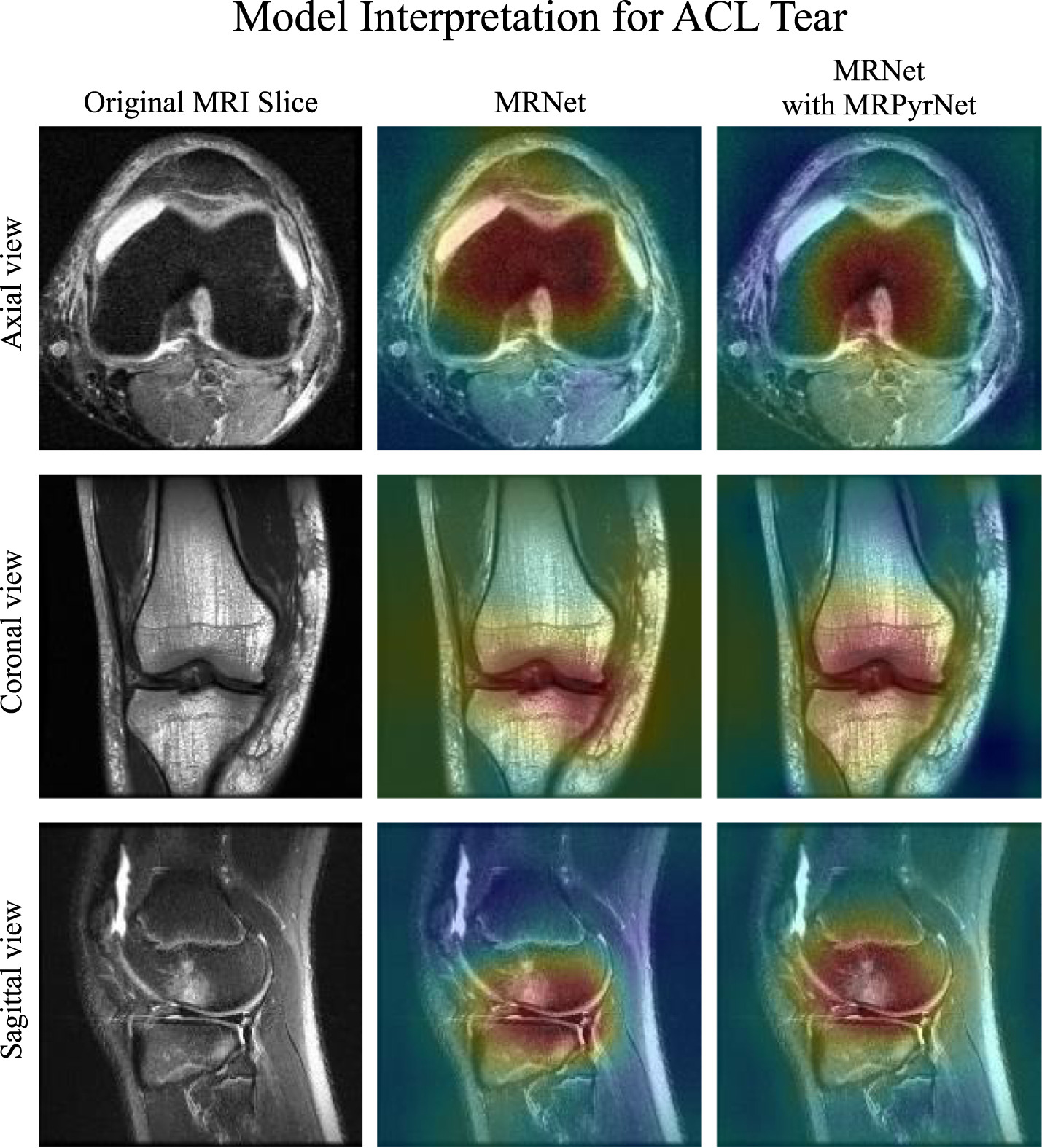}} \par (a) ACL tear
\par \raisebox{-\totalheight}{\includegraphics[width=0.21\textwidth]{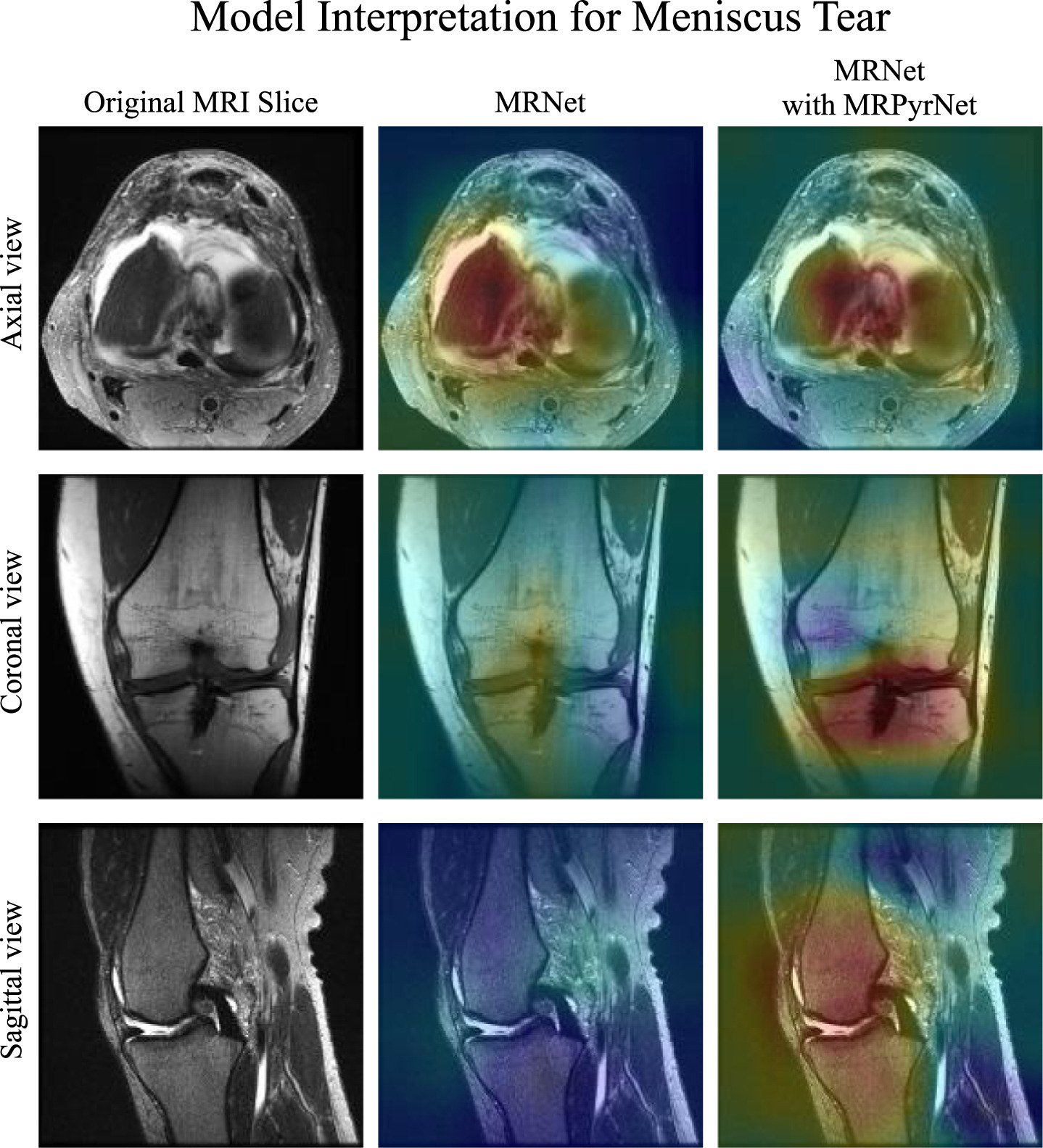}} \par (b) Meniscus tear
\\
\hline

\citet{Wang2023confident}
& GradCAM
& Imaging - X-ray
& Siamese-GAP model with hybrid loss strategy (89.14\% accuracy, 86.78\% F1)
& Presence of early OA \par (2 classes)
& No further explanation by the authors.
& \raisebox{-\totalheight}{\includegraphics[width=0.21\textwidth]{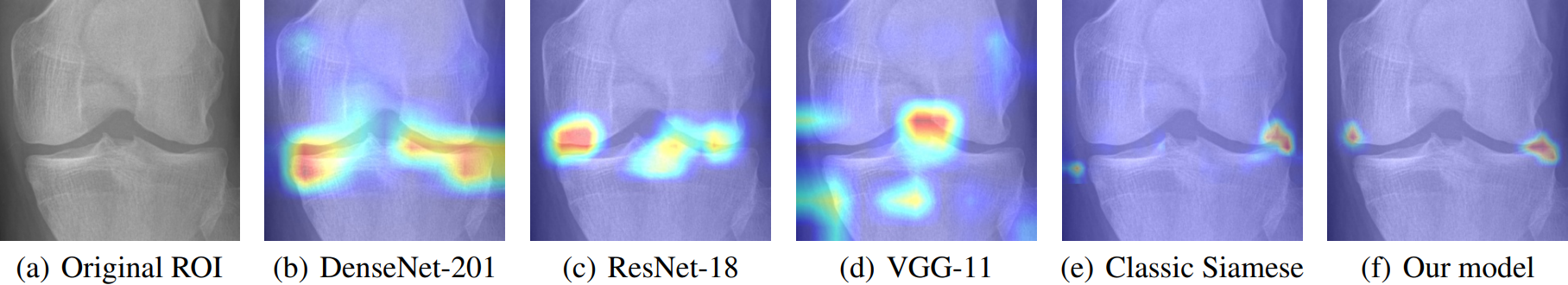}}
\\
\hline

\citet{Tariq2023knee}
& Eigen-CAM
& Imaging - X-ray
& Ensemble CNN (ResNet-34, VGG-19, DenseNet 121, and DenseNet 161) with transfer learning (98\% overall accuracy and 0.99 Quadratic Weighted Kappa)
& Classification of OA severity based on KL grades \par (5 classes)
& The model allowed for the distinction of bone sclerosis, osteophytes, cartilage degeneration, and JSN by highlighting their extracted features. Notably, the Ensemble model achieved nearly 100\% accuracy for KL0, while ResNet-34 and DenseNet-121 exhibited improved feature identification for KL4.
& \raisebox{-\totalheight}{\includegraphics[width=0.21\textwidth]{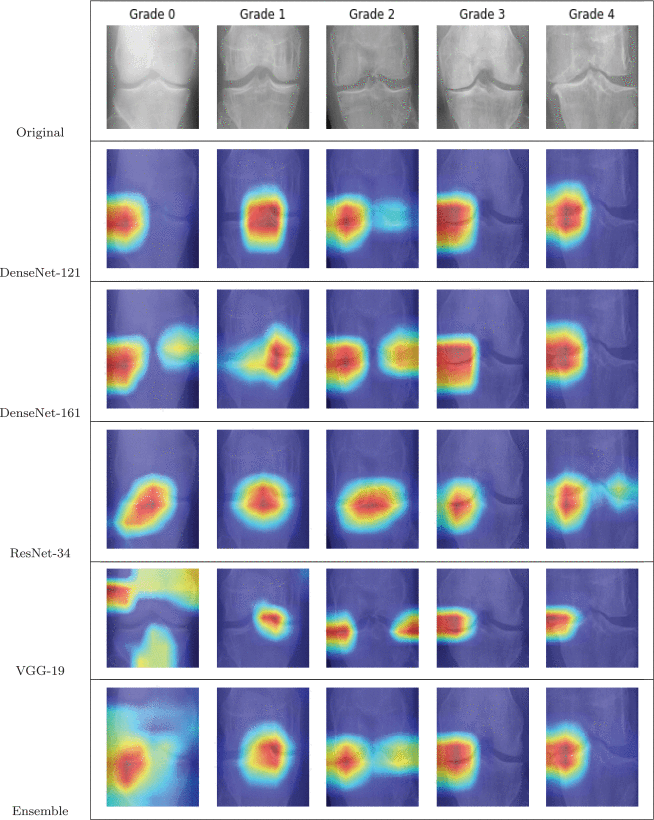}}
\\

\bottomrule
\end{tabularx}}
\end{table*}

\begin{table*}[]
\ContinuedFloat
\caption{(\textit{continued}).}
\resizebox{\textwidth}{!}{%
\begin{tabularx}{\textwidth} {p{1.0cm} p{1.25cm} p{1.5cm} p{2.0cm}  p{1.5cm} p{3.0cm} p{5.25cm}}
\toprule
\textbf{Paper} & \textbf{XAI method} & \textbf{Type of data} & \textbf{Evaluated model (performance)} & \textbf{Target} & \textbf{XAI findings} & \textbf{Visualization of XAI}\\
\midrule

\citet{Li2023deep}
& GradCAM
& Imaging - X-ray
& ResNet50 (0.88 accuracy when using anteroposterior knee X-ray, 0.93 accuracy when using multiview X-ray images)
& Classification of OA severity based on KL grades \par (5 classes)
& Heatmap region was predominantly focused on the knee area, which aligns with expectations. However, in certain images, the study observed that the heatmap region extended beyond the knee joint to other tissues, suggesting that certain extra-articular tissues may also play a key role in knee OA diagnosis.
& \raisebox{-\totalheight}{\includegraphics[width=0.21\textwidth]{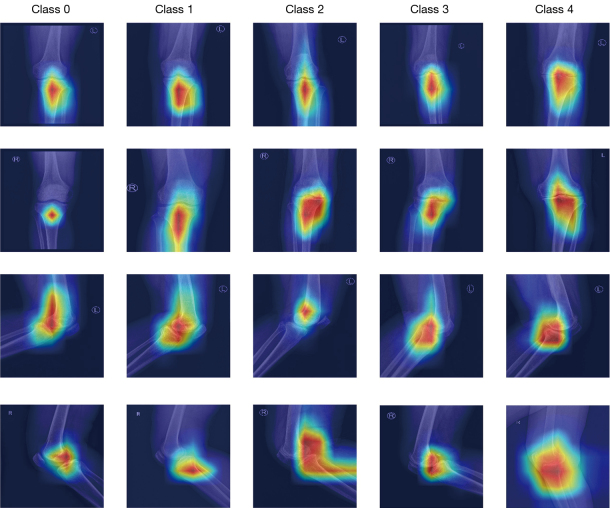}}
\\
\hline

\citet{Wang2023transformer} 
& GradCAM
& Imaging - X-ray
& Vision Transformer (ViT) model with Selective Shuffled Position Embedding (SSPE) and a ROI-exchange strategy (89.80\% accuracy and 87.66\% F1)
& Presence of early OA \par (2 classes)
& All models detected early knee OA features like osteophytes and JSN. However, the DenseNet, ResNet, and VGG models were influenced by background noise, leading to reduced classification performance. The proposed approach and Siamese-based models focused on specific regions affected by knee OA, resulting in better performance.
& \raisebox{-\totalheight}{\includegraphics[width=0.21\textwidth]{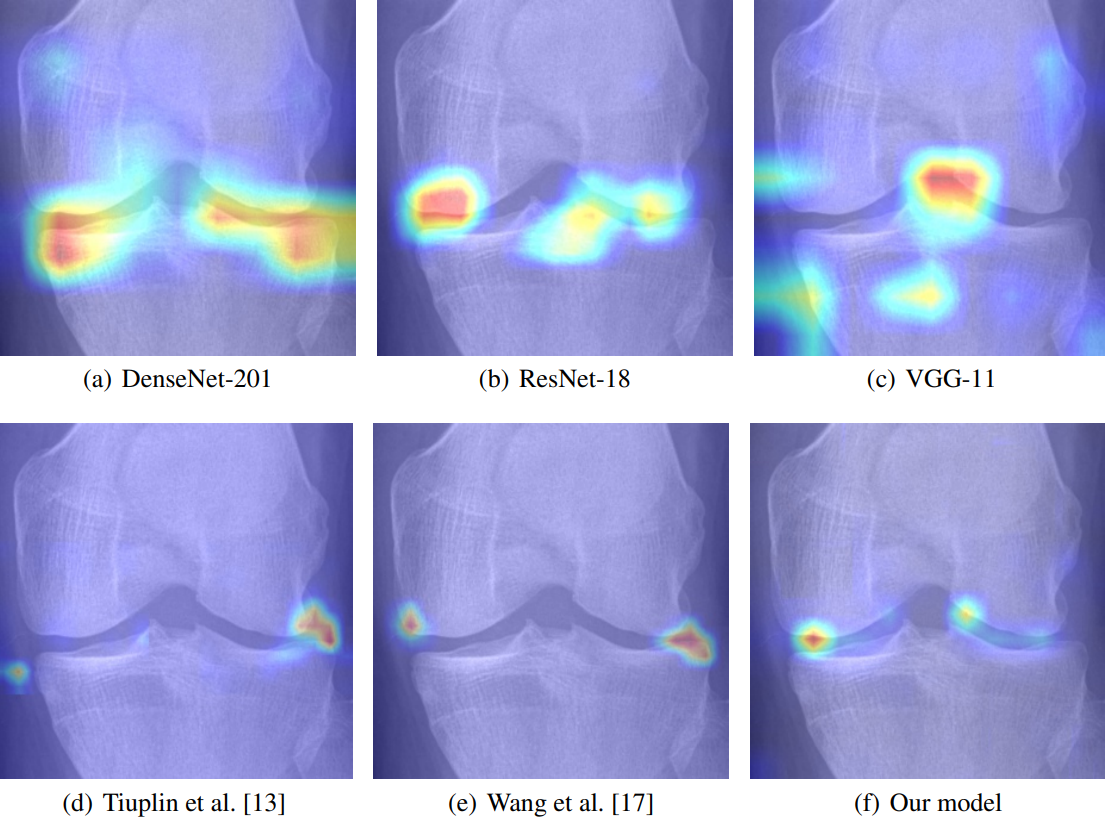}}
\\
\hline

\citet{Farajzadeh2023ijes}
& GradCAM
& Imaging - X-ray
& A deep residual neural network with eight convolutional layers, termed IJES-OA Net (80.23\% average accuracy and 0.802 average precison)
& Classification of OA severity based on KL grades \par (5 classes)
& The proposed model focused on the edges of bones near joint space area.
& \raisebox{-\totalheight}{\includegraphics[width=0.21\textwidth]{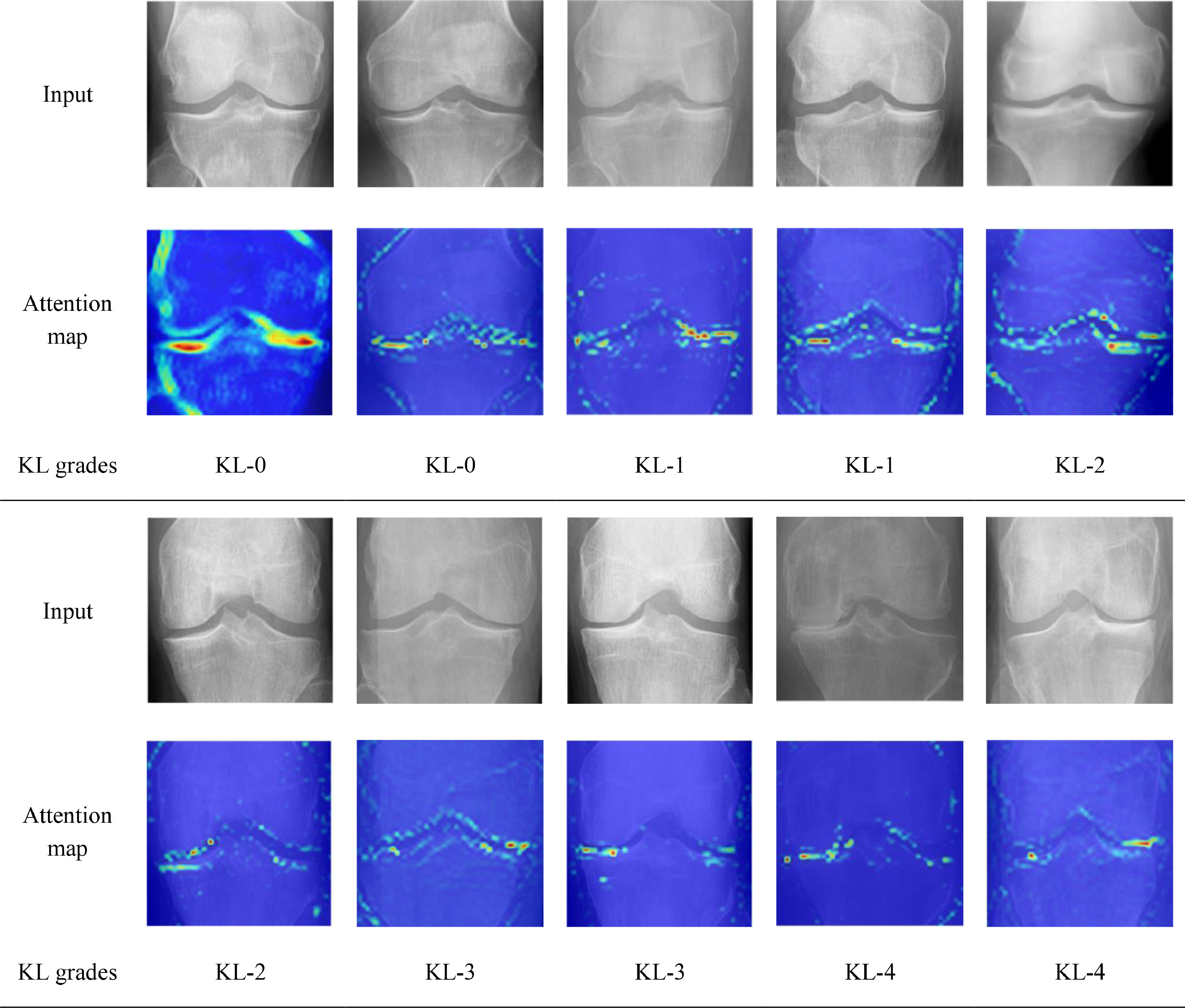}}
\\

\bottomrule
\end{tabularx}}
\end{table*}

\begin{table*}[]
\ContinuedFloat
\caption{(\textit{continued}).}
\resizebox{\textwidth}{!}{%
\begin{tabularx}{\textwidth} {p{1.0cm} p{1.25cm} p{1.5cm} p{2.0cm}  p{1.5cm} p{3.0cm} p{5.25cm}}
\toprule
\textbf{Paper} & \textbf{XAI method} & \textbf{Type of data} & \textbf{Evaluated model (performance)} & \textbf{Target} & \textbf{XAI findings} & \textbf{Visualization of XAI}\\
\midrule

\citet{Wang2023successful}
& Attention maps
& Imaging - X-ray
& CNN (78\% accuracy)
& Classification of OA severity based on KL grades \par (4 classes)
& Heatmap highlighted the area around the knee joint, including the joint space, osteophytes, tibial plateau, and femoral condyles.
& \raisebox{-\totalheight}{\includegraphics[width=0.21\textwidth]{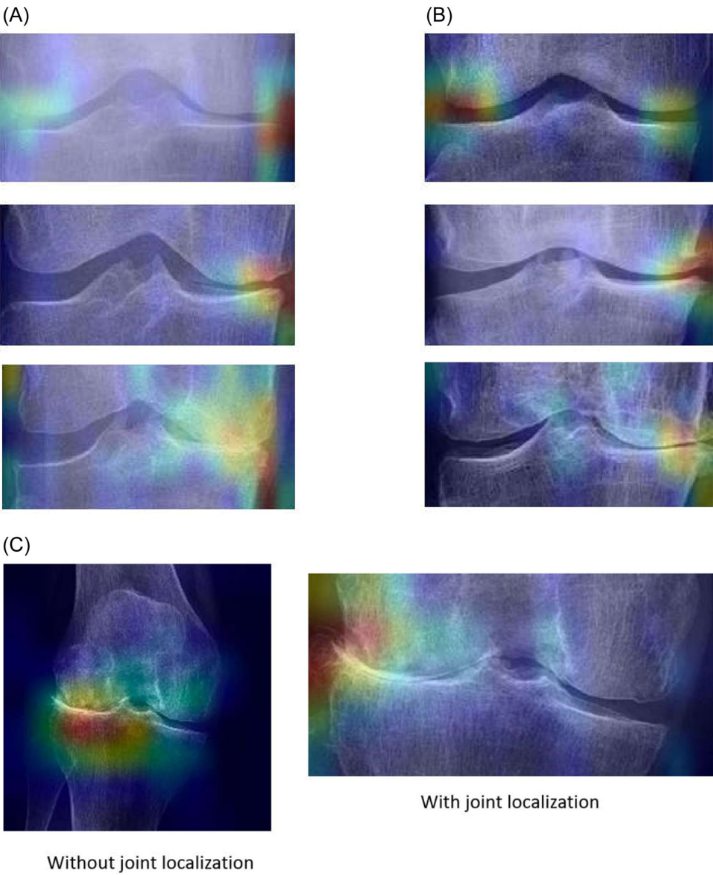}}
\\
\hline

\citet{Tack2021}     
& SmoothGrad saliency maps
& Imaging - MRI
& ResNet (0.84-0.96 AUC)
& Detection of meniscal tears \par (2 classes)
& SmoothGrad saliency maps were found to be effective in highlighting multiple affected sub-regions. The Dilation ResNet-C-26 yields smoother saliency maps, while ResNet-50 may target the affected region better but with less sharp outlines.
& \raisebox{-\totalheight}{\includegraphics[width=0.21\textwidth]{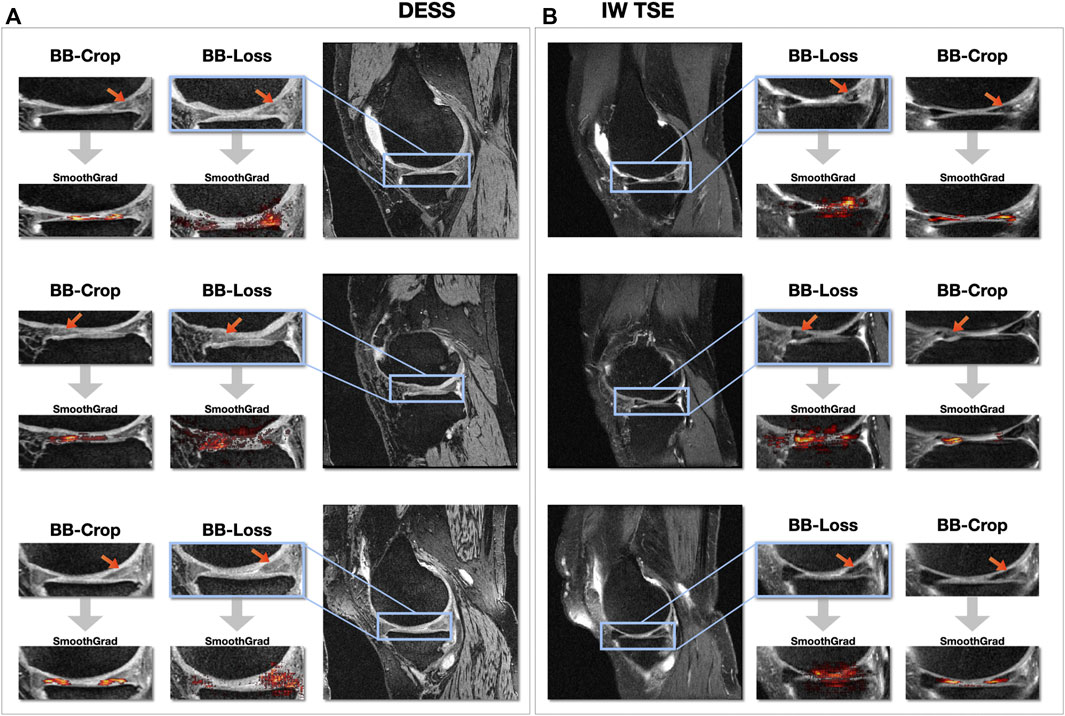}}
\\

\bottomrule
\end{tabularx}}
\end{table*}

\begin{table*}[]
\caption{Summary of game-theory-based XAI techniques from included papers. JSN: joint space narrowing; SHAP: SHapley Additive exPlanations; XGBoost: eXtreme Gradient Boosting.}\label{tab:game_theory_based}
\resizebox{\textwidth}{!}{%
\begin{tabularx} {\textwidth}{p{1.0cm} p{1.0cm} p{1.25cm} p{2.35cm}  p{1.5cm} p{7.0cm} }
\toprule
\textbf{Paper} & \textbf{XAI method} & \textbf{Type of data} & \textbf{Evaluated model (performance)} & \textbf{Target} & \textbf{XAI findings} \\
\midrule

\citet{Kokkotis2020machine}    
& SHAP
& Tabular clinical data
& Logistic regression (77.88\% accuracy with 40 risk factors)
& Presence of OA \par (2 classes)
& Top ten risk factors \par
1. Right knee symptoms: swelling, last 7 days (V00KSXRKN1) \par 
2. Left knee symptoms: bend knee fully, last 7 days (V00KSXLKN5) \par
3. Either knee, history of knee surgery (P02KSURG) \par
4. Knee symptoms, risk factors, or both, status at initial eligible interview or screening visit (P02ELGRISK) \par
5. Baseline symptomatic knee OA status by person (P01SXKOA) \par
6. Right knee exam: patellofemoral crepitus present on exam (V00RKPFCRE) \par
7. Left knee exam: patellofemoral crepitus present on exam (V00LKPFCRE) \par
8. Left knee symptoms: swelling, last 7 days (V00KSXLKN1) \par
9. Average current scale weight in kg (P01WEIGHT) \par
10. Right knee baseline symptomatic OA status (P01RSXKOA)
\\
\hline

\citet{Ntakolia2021identification}
& SHAP
& Tabular clinical data
& Logistic regression (83.3\% accuracy with 29 risk factors)
& Prediction of JSN progression \par (2 classes)
& The three most significant features were identified: lateral JSN on the right knee (P01SVRKJSL), lateral JSN on the left knee (P01SVLKJSL), and measure related to the percentage of foods marked as a small portion (V00PCTSMAL). 
\\
\hline

\citet{Kokkotis2021identifying}
& Kernel SHAP
& Tabular clinical data
& GenWrapper employing SVM (71.25\% mean accuracy with 35 risk factors)
& Prediction of OA progression \par (2 classes)
& The most important variables that significantly influenced the prediction output were identified as lateral JSN on the left knee (P01SVLKOST), body mass index, average daily nutrients from vitamin supplements (V00SUPCA), and education level (V00EDCV). 
\\
\hline

\citet{Ntakolia2021identification}
& SHAP
& Tabular clinical data
& XGBoost (78.14\% average accuracy with 31 risk factors)
& Prediction of JSN progression \par (2 classes)
& Top five risk factors \par
1. Lateral JSN on the right knee (P01SVRKJSL) \par 
2. Percentage of foods marked as large portion (V00PCTLARG) \par
3. Frequency of cream/half and half/non-dairy creamer in coffee or tea in the past 12 months (V00FFQ68) \par
4. Frequency of getting in and out of squatting position 10 or more times during a typical week in the past 30 days (V00PA530CV) \par
5. Lateral JSN on the left knee (P01SVLKOST)
\\
\hline

\citet{Kokkotis2022}     
& SHAP
& Tabular clinical data
& Fuzzy feature selection and random forest (73.55\% accuracy, 73.82\% precision, 73.64\% recall, 73.59 F1 with 21 risk factors)
& Presence of OA \par (2 classes)
& Top five risk factors \par
1. Knee symptoms (P02ELGRISK) \par 
2. History of knee surgery (P02KSURG) \par
3. Age (V00AGE) \par
4. BMI (P01BMI) \par
5. KOOS quality of life score (V00KOOSQOL)
\\

\bottomrule
\end{tabularx}}
\end{table*}

\begin{table*}[]
\ContinuedFloat
\caption{(\textit{continued}).}
\resizebox{\textwidth}{!}{%
\begin{tabularx} {\textwidth}{p{1.0cm} p{1.0cm} p{1.25cm} p{2.35cm} p{1.5cm} p{7.0cm}}
\toprule
\textbf{Paper} & \textbf{XAI method} & \textbf{Type of data} & \textbf{Evaluated model (performance)} & \textbf{Target} & \textbf{XAI findings} \\
\midrule

\citet{Angelini2022}     
& SHAP TreeExplainer
& Biochemical markers
& K-means clustering, random forest or KNN (F1 scores of 0.85 for C1 vs rest, 0.91 for C2 vs rest, and 0.88 for C3 vs rest)
& OA dominant molecular endotypes \par (3 clusters)
& \textbf{Cluster 1 - low tissue turnover:} This cluster demonstrated low repair and articular cartilage or subchondral bone turnover, and had the highest proportion of non-progressors. \par 
\textbf{Cluster 2 - structural damage:} This cluster demonstrated high bone formation/resorption, cartilage degradation and was mostly linked to longitudinal structural progression. \par
\textbf{Cluster 3 - systemic inflammation:} This cluster demonstrated joint tissue degradation, inflammation, and cartilage degradation, and was linked to sustained or progressive pain.
\\
\hline

\citet{Kokkotis2022leveraging}     
& SHAP
& Tabular gait data
& SVM (94.95\% accuracy, 92.16–96.72\% precision, 92.19–97.62\% recall, and 93.07–96.47\% F1 score)
& Classification of anterior cruciate ligament injury status \par (3 classes)
& The gait parameters K2, H4, A3, GRF4, GRF7, K1, A4, and GRF6 as illustrated in below figure were identified as the key factors that significantly influenced the model output, with mean SHAP values higher than 0.3. 
\raisebox{-\totalheight}{\includegraphics[width=0.40\textwidth]{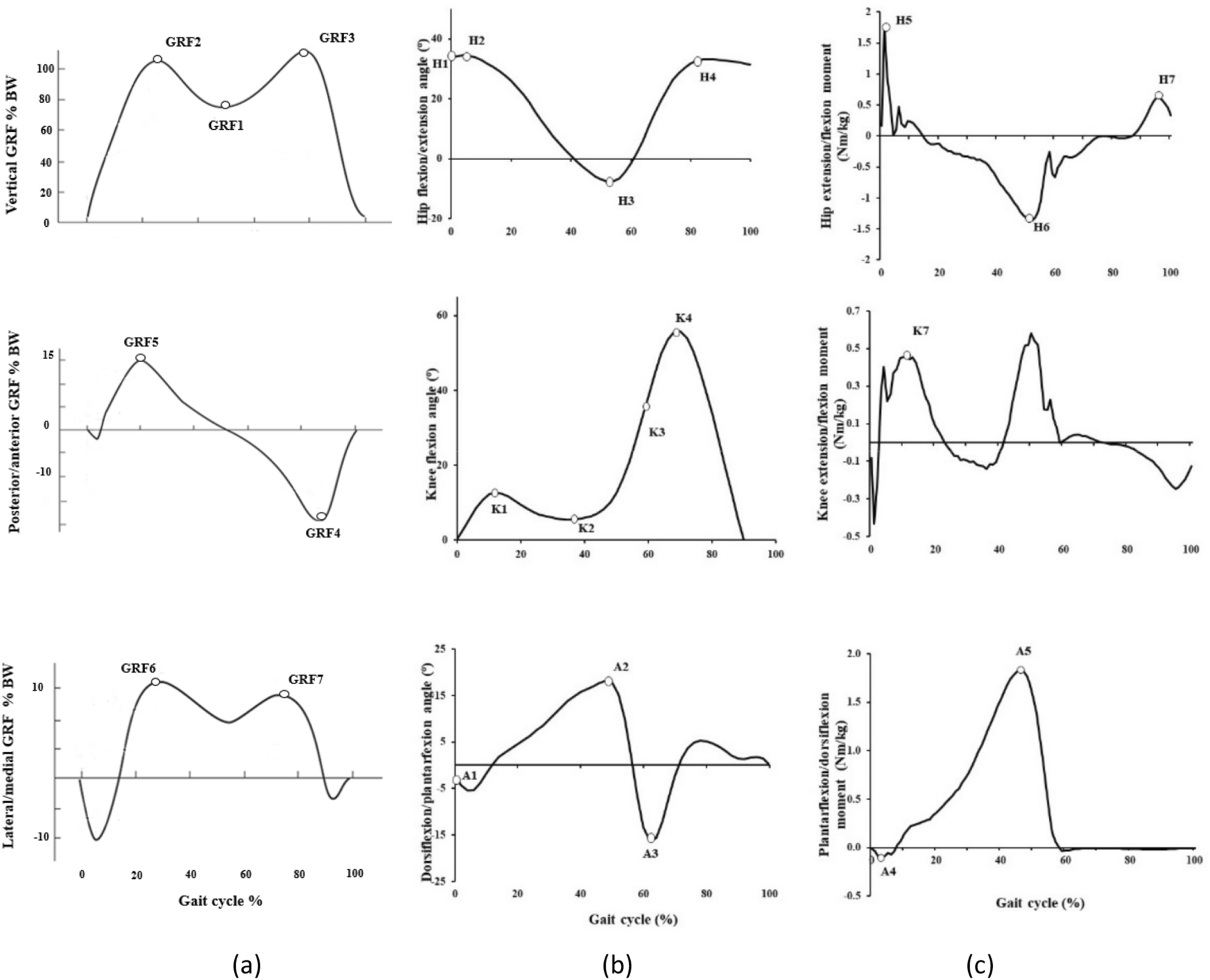}}
\\
\hline

\citet{Lu2022}     
& SHAP
& Tabular clinical data
& XGBoost (0.741 AUC with 15 features)
& Prediction of risk of developing venous thrombosis \par (2 classes)
& KL grade, age, and hypertension emerged as the three pivotal variables in relation to the risk of venous thrombosis
\\

\bottomrule
\end{tabularx}}
\end{table*}

\subsection{Evaluation of XAI}
Evaluation of XAI could be guided by the Co-12 properties introduced by \citet{Nauta2023}. These properties include Correctness, Completeness, Consistency, Continuity, Contrastivity, Covariate complexity, Compactness, Composition, Confidence, Context, Coherence, and Controllability. However, despite the comprehensive framework provided by these properties, our observations indicate that there has been relatively less emphasis on evaluating the explanations generated by XAI in existing research.

Limited evaluation methods for XAI were identified, such as sensitivity analysis \citep{Pierson2021}, confident score \citep{Wang2021learning}, and rate of agreement with medical experts \citep{Chang2020}, which minimally addressed the Correctness and Confidence properties. Nevertheless, it is worth highlighting the work of \citet{Chang2020} who went beyond the general human perception interpretation of GradCAM visualizations. They actively involved medical experts in the validation process to evaluate the reliability of the visualization maps.

Their approach underscores the importance of soliciting feedback and insights from domain experts to evaluate the reliability and effectiveness of XAI techniques. By incorporating the perspectives of medical experts, the evaluation of XAI can benefit from their specialized knowledge and experience, ultimately enhancing the reliability and applicability of the generated explanations.

\section{Discussion}
\label{sec:discussion}
XAI holds promise in the identification of pathological patterns associated with knee OA by leveraging structured and unstructured data from diverse sources. Moreover, it has the potential to optimize the handling and organization of electronic health record data, resulting in streamlined clinical workflows and significantly reducing the time physicians spend on making diagnosis, prognosis, and searching for pertinent patient information in electronic records. However, the current state of XAI has certain limitations that need to be addressed in future research.

\subsection{Development of quantitative evaluation metrics for XAI}
Our findings shed light on a significant gap in the evaluation of XAI generated explanations, where qualitative assessments are predominantly utilized. However, these subjective evaluations may not meet the evidence standards required by medical experts. Hence, the development of a quantitative evaluation metric for XAI is essential. Such a metric can provide objective measures and benchmarks for comparing different XAI techniques and their impact on decision-making processes. It can also help researchers and practitioners in the field to determine the strengths and weaknesses of their models, identify areas for improvement, and facilitate the reproducibility of results.

\subsection{Integration of domain-specific information from stakeholders}
The deployment of XAI could lead to the real application of AI in healthcare, and overcome the lack of operator confidence in AI models. However, it is essential to understand how the application of these models in clinical tasks will be perceived, whether as a support or a substitute for medical expert's work, as well as the level of substitution. To achieve this, AI programmers must discern which explanations are valuable and which are not for medical professionals. Creating an XAI model that is deemed useless or difficult to comprehend may deter medical experts from utilizing it. Furthermore, patients play a significant role as stakeholders since the developed model aims to elucidate their health status. Therefore, their expectations and special needs should be taken into account and integrated into the process. To address the challenges, \citet{Mrklas2020} implemented a qualitative co-design approach at an academic health center in Southern Alberta, which involved conducting focus groups with patients, physicians, researchers, and industry partners, as well as analyzing prioritization activities and a pre-post quality and satisfaction Kano survey. The structured co-design processes were developed based on the basis of shared concepts, language, power dynamics, rationale, mutual learning, and respect for diversity and differing opinions.

\subsection{Exploring patient disparities in data and addressing population-specific factors}
As highlighted by \citet{Pierson2021}, there are noticeable racial and socioeconomic disparities in OA data. By considering these disparities during the training of AI models, there is a potential to enhance accuracy. The study also revealed that patient-perceived OA symptoms vary based on factors such as education, culture, and geography. Considering these variations is crucial in developing AI models that accurately capture the diverse experiences and manifestations of OA among different patient populations. In addition, we observed that there is lack of well-organized open access data specifically for the Eastern population, despite the higher prevalence of OA issues in this population \citep{Inoue2001prevalence}. This highlights a significant gap in available resources for studying and addressing OA within the Eastern population. The limited availability of comprehensive and representative data from this specific demographic group hinders the development and evaluation of AI models tailored to their unique needs and characteristics. 

\subsection{Understanding the boundaries of knowledge and legal constraints}
It is crucial to recognize that XAI is not all-powerful. Identifying the knowledge boundaries of XAI models is upmost important for users to have a clear understanding and make appropriate use of the tool. Users should be informed about the operational boundaries of the models and be able to discern when the models go beyond their knowledge limits, as this can potentially result in errors. While XAI-based systems have the potential to alleviate the workload of healthcare providers, they also raise concerns regarding legal responsibility in cases of unethical actions and errors. Therefore, the development of XAI models in healthcare should be approached with caution, while also recognizing their potential to bring about positive societal impacts.

\subsection{Exploring alternative XAI techniques for knee OA applications}
In addition to the XAI applications discussed in the Section \ref{sec:discussion}, a prospective XAI technique in OA diagnosis could be image captioning. It is a process of generating a textual description of an image using AI algorithms. Medical imaging is an area where this technology could be particularly useful, as generating accurate and detailed descriptions of radiology and pathology images could help healthcare professionals to identify the specific areas of the knee that require treatment and make better-informed decisions about patient care. This area of research presents an exciting opportunity for the development of new XAI models that could have a significant impact on the future of musculoskeletal healthcare. Furthermore, the exploration of the counterfactual approach to XAI in the context of OA applications presents an additional avenue for research. This approach aims to enhance people understanding of AI systems by offering counterfactual explanations specific to target domain. Recent studies have shown that counterfactuals can provide richer information compared to causal explanations, as they encompass a broader range of possibilities in their mental representation \citep{Celar2023people}.  

\section{Conclusion}
\label{sec:conclusion}
A substantial number of studies in the field of computer-aided diagnosis for knee OA have sought to incorporate explainability through various XAI techniques in their deep learning models. However, these techniques encounter inherent limitations due to the absence of a robust XAI framework, and the evaluation of explanation quality is often overlooked, leading to uncertainty about the effectiveness of these approaches in addressing the black box nature of deep learning in OA diagnosis. Nevertheless, the development of XAI in knee OA detection aligns with the trend of precision diagnosis, offering the potential to reduce the healthcare burden and promote preventive strategies for musculoskeletal diseases.

\section*{Acknowledgment}
This work was supported by Ministry of Higher Education, Malaysia under Fundamental Research Grant Scheme (FRGS) Grant No. FRGS/1/2022/SKK01/UM/02/1.

\printcredits

\clearpage
\clearpage
\bibliographystyle{cas-model2-names}

\bibliography{cas-refs}

\end{document}